\documentclass{article}

\usepackage{microtype}
\usepackage{graphicx}
\usepackage{subcaption}
\usepackage{booktabs} 
\usepackage{longtable}
\usepackage{xcolor}
\usepackage{titletoc}
\usepackage[colorlinks,linkcolor=blue]{hyperref}

\usepackage{hyperref}

\usepackage[accepted]{icml2026}

\usepackage{amsmath}
\usepackage{amssymb}
\usepackage{mathtools}
\usepackage{amsthm}
\usepackage{titletoc}
\usepackage{multirow}
\usepackage{tabularx}
\usepackage{xcolor}
\usepackage[capitalize,noabbrev]{cleveref}

\theoremstyle{plain}

\theoremstyle{definition}

\theoremstyle{remark}

\usepackage[textsize=tiny]{todonotes}

\icmltitlerunning{Training-Free Acceleration of dLLMs via Draft-Guided Contiguous Leaping Decoding}

\newcommand{\AppTOCTitle}{\section*{Contents}}

\newcommand{\apptocA}[2]{%
  \noindent\hyperref[#1]{\textbf{#2}}%
  \leaders\hbox{\kern.35em.\kern.35em}\hfill%
  \pageref{#1}\par
}

\newcommand{\apptocB}[2]{%
  \noindent\hspace*{1.6em}\hyperref[#1]{#2}%
  \leaders\hbox{\kern.35em.\kern.35em}\hfill%
  \pageref{#1}\par
}

\begin{document}

\twocolumn[
  \icmltitle{DC-Leap: Training-Free Acceleration of dLLMs \\ via Draft-Guided Contiguous Leaping Decoding}
  \icmlsetsymbol{corres}{$\dagger$}
  \begin{icmlauthorlist}
    \icmlauthor{Yanhua Jiao$^{1,2}$}{}\quad
    \icmlauthor{Tianyi Wu$^1$}{}\quad
    \icmlauthor{Xiaoxi Sun$^1$}{}\quad
    \icmlauthor{Yulin Li$^1$}{} \\
    \icmlauthor{Huiling Zhen$^3$}{}\quad
    \icmlauthor{Libo Qin$^1$}{}\quad
    \icmlauthor{Baotian Hu$^{1,2}$}{}\quad
    \icmlauthor{Zhuotao Tian$^{1,2}$}{corres} \quad
    \icmlauthor{Min Zhang$^{1,2}$}{}
  \end{icmlauthorlist}
  \begin{center}
    \textsuperscript{1}Harbin Institute of Technology, Shenzhen \quad
    \textsuperscript{2}Shenzhen Loop Area Institute  \quad
    \textsuperscript{3} Huawei Noah’s Ark Lab
  \end{center}
  \icmlcorrespondingauthor{$^\dagger$ Zhuotao Tian}{tianzhuotao@hit.edu.cn}
  \icmlkeywords{Machine Learning, ICML}
  \vskip 0.3in
  
]
\makeatletter
\global\icml@noticeprintedtrue
\insert\footins{
  \vspace{5pt}
  \footnotesize
  \noindent
  \textsuperscript{$\dagger$}Corresponding Author: Zhuotao Tian \\ $<$tianzhuotao@hit.edu.cn$>$.
  
  \vspace{2pt}
  \noindent \textit{Proceedings of the $43^{rd}$ International Conference on Machine Learning}, Seoul, South Korea. PMLR 306, 2026. Copyright 2026 by the author(s).
}
\makeatother

\begin{abstract}
  While parallel decoding is central to the efficiency of Diffusion Large Language Models (dLLMs), current strategies are often hindered by overly conservative confidence thresholds. These thresholds, necessitated by the Joint Probability Dependence Error (JPDE), result in redundant denoising iterations and suboptimal inference speeds.
  To overcome this, we propose DC-Leap, a training-free framework that enables reliable acceleration of dLLMs in the moderate-confidence regime. DC-Leap introduces a Dynamic Contiguous Verification strategy that integrates strictly-ordered causal constraints into the parallel decoding process. By progressively validating token dependencies, this mechanism effectively neutralizes the JPDE, enabling reliable acceleration with comparable performance. Furthermore, DC-Leap incorporates the draft-guided decoding mechanism, where the draft helps extend the context by leaping forward across multiple tokens, providing look-ahead context and retaining the structural benefits of bidirectional attention during inference.
  Extensive experiments on standard benchmarks demonstrate that DC-Leap achieves substantial speedups, 
  up to \textbf{53.19$\times$} on MBPP for long-sequence generation, and up to \textbf{105.02$\times$} when combined with KV-Cache with comparable generation quality. Code is available at \url{https://github.com/ffh-wyls/DC-Leap}.
\end{abstract}

\begin{figure}[h]
    \centering
    \includegraphics[width=0.95\linewidth]{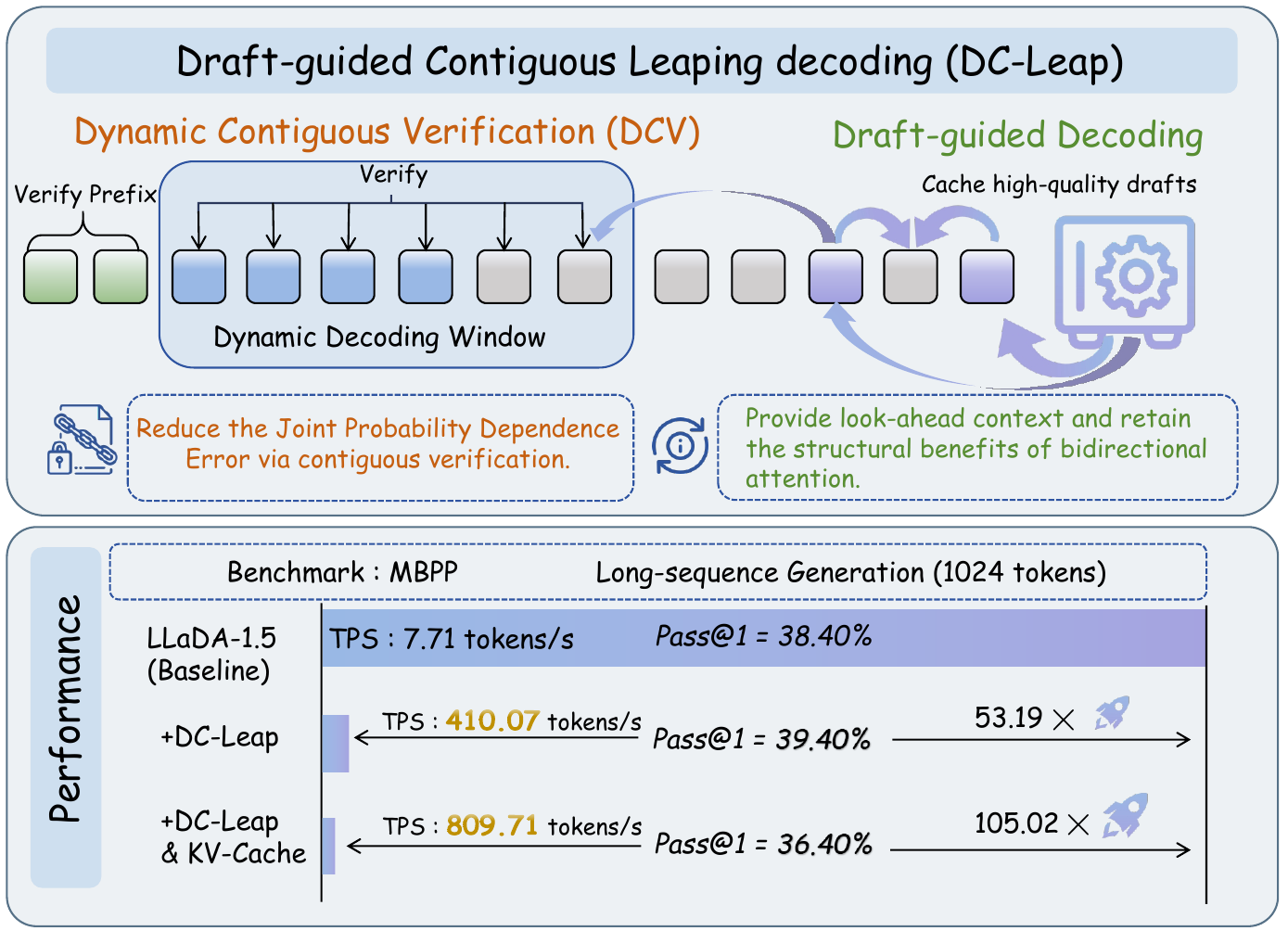}
    \caption{\textbf{Overview and Performance of DC-Leap.} Top: Schematic of the proposed training-free framework, featuring Dynamic Contiguous Verification to neutralize JPDE through causal constraints, and a Draft Mechanism to expedite convergence via distal context. Bottom: Evaluation on MBPP for long-sequence generation. DC-Leap achieves up to \textbf{53.19$\times$} significant speedup while maintaining near-lossless performance.}
    \label{fig:teaser}
\end{figure}

\section{Introduction}
\label{sec:intro}
Diffusion Large Language Models (dLLMs) such as LLaDA ~\citep{llada, llada1.5} and Dream~\citep{dream7b}, have emerged as a compelling generative paradigm, challenging the dominance of Autoregressive (AR) models by offering non-autoregressive capabilities and achieving generation quality comparable to models of similar scale. Unlike the strict sequential dependency inherent in AR architectures, dLLMs inherently support parallel decoding, presenting a viable path toward overcoming the 
latency bottlenecks associated with sequential token generation. However, dLLMs may often 
resort to redundant iterative refinement, resulting in inference throughput that trails behind highly optimized AR models, thereby limiting the utility of dLLMs in latency-sensitive applications.

\begin{figure*}[t]
    \centering
    \begin{subfigure}[t]{0.48\textwidth}
        \vspace{0pt}
        \centering
        \includegraphics[width=0.85\textwidth]{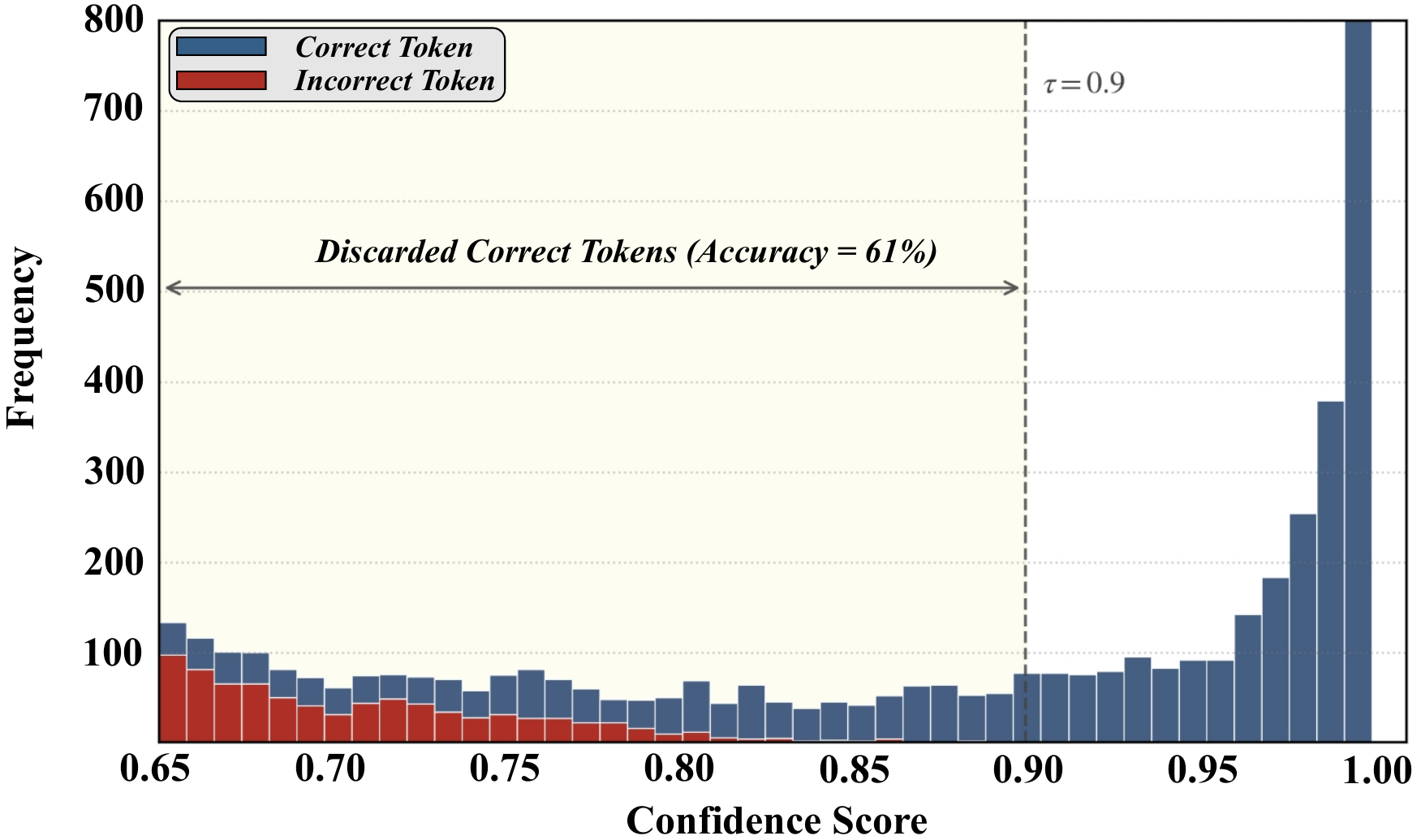} 
        \caption{The frequency of correct versus incorrect predictions across different confidence intervals from LLaDA-1.5. Tokens in the loose confidence range ($0.65\text{-}0.9$) account for a significant portion of correct predictions yet are discarded by conservative thresholds.}
        \label{fig:threshold_cost}
    \end{subfigure}
    \hfill
    \begin{subfigure}[t]{0.48\textwidth}
        \vspace{0pt}
        \centering
        \includegraphics[width=0.85\textwidth]{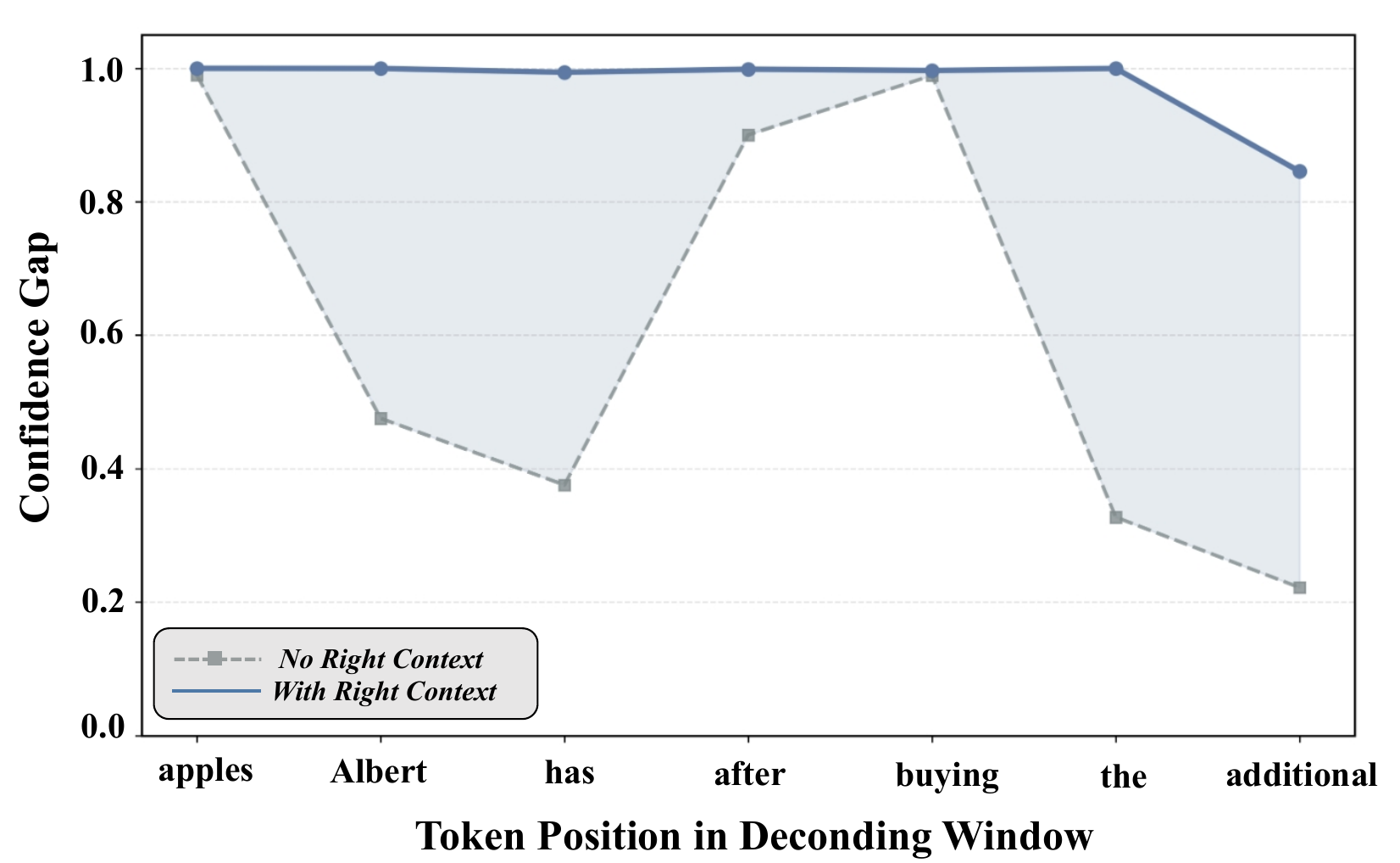} 
        \caption{Confidence gap ($P_{\text{top1}} - P_{\text{top2}}$) for intermediate tokens. The presence of right-context (blue line) significantly widens the margin relative to the baseline (grey line). This trend demonstrates that future information sharpens the predictive distribution, thereby reducing ambiguity during the generation of intermediate tokens.}   
        \label{fig:future_quality_impact}
    \end{subfigure}
    
    \caption{\textbf{Motivation of DC-Leap.} (a) Strict confidence thresholds in existing parallel decoding methods lead to the waste of potentially correct tokens. (b) Leveraging right-side context reduces the confidence gap, facilitating faster convergence.}
    \label{fig:intro_motivation}
\end{figure*}

\paragraph{Motivation.}
To mitigate dLLM's computational burden, recent research has focused on parallel decoding strategies, such as Fast-dLLM~\cite{fast-dllm} and LocalLeap~\cite{local-leap}. These methods accelerate decoding by accepting multiple tokens per step based on the confidence thresholds. Typically, these methods use strict, high-confidence thresholds, which are usually above 0.9. 
However, as depicted in Figure~\ref{fig:intro_motivation} (a), a substantial fraction of correct tokens (61\% within the 0.65--0.9 confidence interval) are often discarded due to falling below the predefined confidence threshold. This premature filtering can create an efficiency bottleneck by restricting the model's capacity to leverage rapid token convergence in lower-confidence regions, thus limiting the potential acceleration gains.

In fact, such stringent thresholding primarily arises from the need to manage errors introduced by parallel acceptance strategies: their effectiveness depends on the assumption that \textit{token predictions are conditionally independent given the current context}. However, recent findings by~\citet{EB-Sampler} highlight that parallel decoding inherently induces the \textit{Joint Probability Dependence Error (JPDE)}, caused by simultaneously predicting multiple tokens while neglecting their mutual dependencies, resulting in incoherent outputs.
Thus, a key research question emerges: \textit{Can we mitigate the dependence-induced errors, i.e., JPDE,  through alternative methods, beyond stringent confidence thresholding, allowing for safely reduced thresholds and consequently greater acceleration?}

\paragraph{Limitations of Sequential Decoding.}
A natural approach to mitigating JPDE is to adopt sequential decoding, where tokens are committed one-by-one, conditioning each prediction explicitly on previously accepted tokens. 
In principle, this strategy resolves dependence errors by avoiding simultaneous token predictions within the same iteration; instead, joint predictions across tokens are factorized through the chain rule. However, sequential decoding faces two fundamental limitations in the context of dLLMs.

First, it suffers from \textit{prohibitive inefficiency}: sequential commitment reduces parallelism, causing the number of forward passes, and consequently latency, to scale linearly with sequence length and severely undermining the acceleration effects.

Second, sequential decoding yields an \textit{unfavorable inference regime for bidirectional dLLMs}: the structured, typically left-to-right conditioning pattern inherent to sequential prediction deviates from the distribution encountered during the bidirectional denoising training of dLLMs, as evidenced by the minor confidence gap for most tokens in the No Right Context (fully masked) setting in Figure~\ref{fig:intro_motivation} (b).

Such a mismatch limits the exploitation of bidirectional context at inference, as well as the performance gains brought by bidirectional attention mechanisms.
 
\paragraph{Our Solution.}
To mitigate JPDE while preserving the efficiency of parallel decoding and the advantages of bidirectional context in dLLMs, we introduce \textbf{DC-Leap}, a training-free acceleration strategy via \textbf{D}raft-guided \textbf{C}ontiguous \textbf{Leap}ing decoding. DC-Leap draws inspiration from sequential commitment but circumvents its two major limitations. It consists of two primary mechanisms: 1) \textit{Dynamic Contiguous Verification (DCV)} and 2) \textit{Draft-guided Decoding}.

Specifically, DCV introduces the \textit{dynamic decoding window} that enforces a local, sequential left-to-right verification of tokens. Within each window, tokens are accepted contiguously until a prediction falls below the confidence threshold, thereby replacing the global conditional independence assumption inherent to parallel decoding with a local, window-level factorization, enabling multi-token acceptance while still retaining the efficiency.

On the other hand, to address the distributional mismatch and context limitations caused by the sequential commitment, DC-Leap maintains non-committed \textit{drafts}, i.e., high-confidence tokens predicted outside the decoding window. These drafts help extend the context by \textit{leaping forward} across multiple tokens, providing look-ahead context and retaining the structural benefits of bidirectional attention during inference.

Extensive experiments conducted on three dLLMs across five benchmarks, including math reasoning, instruction following, and code generation, demonstrate the effectiveness
and the generalization capabilities of DC-Leap. 
Our primary contributions can be summarized as follows:
\begin{itemize}
    \item We identify a key trade-off in dLLM acceleration: controlling Joint Probability Dependence Error (JPDE) necessitates strict thresholds in parallel decoding, whereas sequential decoding reduces JPDE at the expense of efficiency and bidirectional inference.
    
    \item We propose \textbf{{DC-Leap}}, an effective decoding framework that integrates dynamic continuous verification with a look-ahead draft mechanism, safely lowering the acceptance threshold for parallel decoding while exploiting bidirectional attention during inference.
    
    \item Extensive experiments on LLaDA-1.5, LLaDA-8B-Instruct, and Dream-v0-7B-Instruct models across five benchmarks validate the effectiveness of DC-Leap, demonstrating substantial inference speedups without compromising the generation quality of the baselines.
\end{itemize}     
\section{Background and Motivation}
\label{sec:background_and_motivation}
In this section, we introduce the preliminaries of dLLMs in Sec.\ref{sec:preliminaries} and present our empirical findings in Sec.\ref{sec:key_observations}, which motivate the designs of our proposed framework.

\subsection{Preliminaries}
\label{sec:preliminaries}

\paragraph{Parallel Decoding in dLLMs.}
Consider a generation task aiming to produce a target sequence $\boldsymbol{x}_{\texttt{gen}}$ of length $M$, conditioned on a prompt $\boldsymbol{x}_{\texttt{prompt}}$ of length $N$. The complete sequence is denoted by $\boldsymbol{x} \in \mathcal{V}^{L}$, with $L = N + M$ and vocabulary $\mathcal{V}$. Let $\left [ \mathrm{MASK}  \right ] \in \mathcal{V}$ represent the mask token.

Unlike autoregressive methods, dLLMs~\cite{llada} employ a non-autoregressive iterative denoising process over a discrete time horizon $t \in \{T, T-1, \dots, 0\}$, which typically starts at step $T$ with a fully masked response appended to the prompt:
\begin{equation}
    \boldsymbol{x}^T = (\boldsymbol{x}_{\texttt{prompt}}, \underbrace{\left [ \mathrm{MASK}  \right ] , \dots, \left [ \mathrm{MASK}  \right ] }_{M \text{ times}}).
\end{equation}
At each denoising iteration $t$, the model $p_\theta$ simultaneously estimates the categorical distribution over the vocabulary for all masked positions. Specifically, for any position $i$ where $\boldsymbol{x}_i^t = \left [ \mathrm{MASK}  \right ] $, the model computes token probabilities:
\begin{equation}
    P(\boldsymbol{x}_i = v \mid \boldsymbol{x}^t) = \left[ \text{Softmax}(p_\theta(\boldsymbol{x}^t)_i) \right]_v, \quad \forall v \in \mathcal{V}.
\end{equation}
Parallel decoding methods for dLLMs~\cite{fast-dllm, EB-Sampler} typically employ a confidence-based thresholding strategy to transition from $\boldsymbol{x}^t$ to $\boldsymbol{x}^{t-1}$. Specifically, for each masked token position $i$, a binary acceptance decision $M_i$ is made independently as follows:
\begin{equation}
    \label{eq:threshold_pd}
    M_i = \mathbb{I}\left( \max_{v \in \mathcal{V}} P(x_i = v \mid \mathbf{x}^t) > \tau \right),
\end{equation}
where $\tau \in [0, 1]$ represents the pre-defined confidence threshold. 
Tokens satisfying $M_i = 1$ are committed simultaneously in a single iteration. 

\paragraph{Joint Probability Dependence Error in Parallel Decoding.}
\label{sec:joint_error}

Parallel decoding strategies accelerate inference by simultaneously predicting multiple masked tokens indexed by the set $U$. These methods approximate the true joint distribution $P(\mathbf{x}_U \mid \mathcal{C})$ with the product of marginals
$Q(\mathbf{x}_U \mid \mathcal{C}) = \prod_{i \in U} p_\theta(x_i \mid \mathcal{C})$, where $C$ is the context that has already been decoded.

This conditional independence assumption introduces the Joint Probability Dependence Error ($\mathcal{E}$). 
As derived in \cite{EB-Sampler}, this error can be quantified by the KL divergence, which is strictly bounded above by the marginal entropies of the target tokens:
\begin{equation}
\mathcal{E} = D_{\text{KL}}(P \parallel Q) \le \sum_{i \in U} H(x_i \mid \mathcal{C}) - \max_{j \in U} H(x_j \mid \mathcal{C}).
\label{eq:entropy_bound}
\end{equation}
For non-contiguous decoding, the sparse context $\mathcal{C}_0$ composed of scattered verified tokens yields high accumulated entropies and loose error bounds. Consequently, these methods compulsorily require conservative confidence thresholds $\tau$ in Eq.~\eqref{eq:threshold_pd} to filter unreliable predictions, severely diminishing their acceleration potential.

\subsection{Key Observations}
\label{sec:key_observations}
As discussed in the Introduction, parallel decoding is fundamentally constrained by JPDE, thus relying on strict confidence thresholds to minimize errors. However, this conservatism leads to excessive rejection of correct tokens in early iterations (Figure~\ref{fig:intro_motivation}~(a)). This raises the core challenge: \textit{how can we safely lower the threshold to recover these tokens without compromising accuracy?}

The problem can be equivalently reformulated as mitigating the JPDE discussed in Sec.~\ref{sec:joint_error} without relying on stringent thresholds. We draw inspiration from sequential decoding, which is theoretically free from JPDE as it strictly follows the probability chain rule: 
\begin{equation}
    P(\mathbf{x}_{1:T} \mid \mathcal{C}) = \prod_{t=1}^{T} P(x_t \mid \mathbf{x}_{<t}, \mathcal{C}).
    \label{eq:sequential_chain}
\end{equation}
where $\mathbf{x}_{1:T}$ represents the target generation sequence of total length $T$, and $\mathcal{C}$ denotes the given context. The $x_t$ refers to the specific token generated at the current time step $t$, which is explicitly conditioned on the fully verified prefix $\mathbf{x}_{<t} = \{x_1, \dots, x_{t-1}\}$ accumulated from previous steps. This ensures that the independence assumption is never violated, thus guaranteeing the generation quality.

\paragraph{Windowed Contiguous Decoding.}
\label{sec:obs_cost}
Although sequential decoding avoids JPDE by construction, its token-by-token commitment is prohibitively inefficient. This motivates a natural question: \textit{Can we retain left-to-right sequentiality while decoding multiple tokens per iteration?}

A promising compromise involves relaxing strict sequentiality to the span level. Consider decoding a candidate span of tokens in parallel, while adhering to a left-to-right commitment rule: only the longest verified contiguous prefix is accepted, meaning that a token is retained only if all preceding tokens in the span are verified. Specifically, as shown in Table~\ref{tab:motivation}, contiguous decoding boosts accuracy compared to non-contiguous alternatives under the same confidence thresholds. This demonstrates that strict sequentiality can coexist with multi-token commitment, effectively reconciling orderly verification with computational efficiency. 

\begin{table}[t]
\centering
\caption{\textbf{Accuracy (\%) on GSM8K in the LLaDA-1.5 under different confidence thresholds $\tau$ for two decoding strategies.}}
\label{tab:motivation}
\renewcommand{\arraystretch}{1.0} 
\setlength{\tabcolsep}{10pt}
\begin{tabular}{@{}lcccc@{}} 
\toprule
Threshold $\tau$ & 0.60 & 0.65 & 0.70 & 0.75 \\
\midrule
Contiguous  & \textbf{77.63} & \textbf{78.85} & \textbf{79.53} & \textbf{80.59} \\
Non-Contiguous  & 75.13 & 77.48 & 79.08 & 80.44 \\
\bottomrule
\end{tabular}
\end{table}

\paragraph{Future Context Benefits.}
Sequential verification enforces a left-to-right commitment pattern, which leaves most future positions masked during decoding. As a result, dLLMs cannot fully exploit their bidirectional attention, since informative right context is largely unavailable at inference time. This motivates a natural question: \textit{Can supplementing reliable right-side future context improve the model's confidence and stability under verification?}

To validate this, we conduct a simple pilot study that provides additional right-context information during decoding and quantifies its effect on prediction uncertainty. Specifically, we measure the confidence gap $P_{\text{top1}} - P_{\text{top2}}$ under varying right-context conditions, where $P_{\text{top1}}$ and $P_{\text{top2}}$ denote the probabilities of the most likely and second-most likely tokens, respectively. As shown in Figure~\ref{fig:intro_motivation} (b), supplying high-quality right context consistently sharpens the confidence gap compared to decoding without right context, indicating more decisive token predictions. 

This suggests that future tokens can serve as semantic anchors that reduce uncertainty and stabilize verification. Moreover, since higher confidence is typically correlated with a higher likelihood of correctness~\cite{guo2017calibration}, confidence provides a practical proxy for selecting useful future cues. These observations motivate our draft mechanism, which retains high-confidence future predictions as look-ahead placeholders to guide decoding.

\section{Methodology}
\label{sec:Methodology}

\begin{figure}[t]
    \centering
    \includegraphics[width=\linewidth]{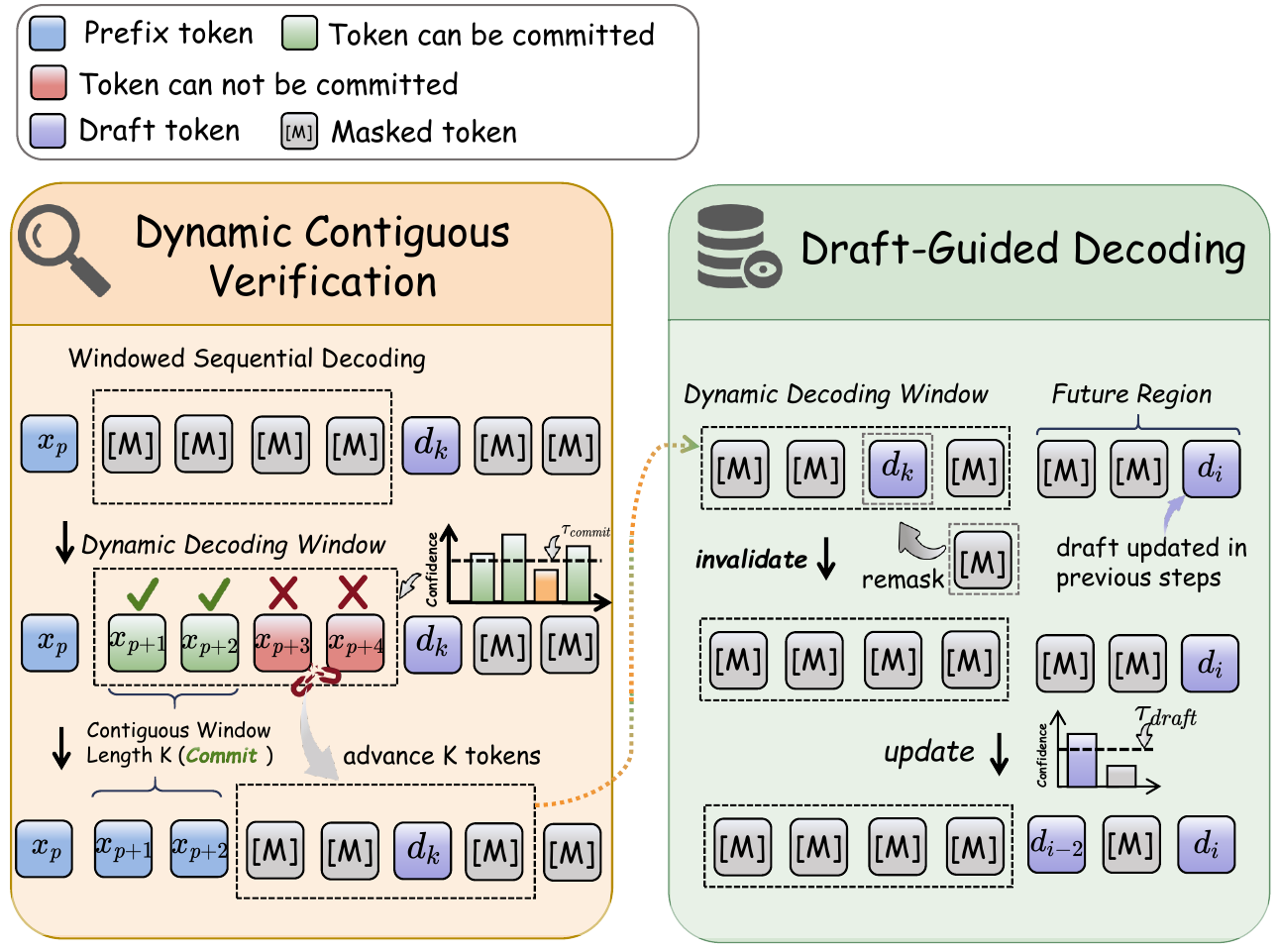}
    \caption{\textbf{Dynamic Contiguous Verification (DCV) and Draft Mechanisms.} Left: DCV determines window length $K$  (tokens exceeding  $\tau_{\text{commit}}$) and advances by committing $K$ contiguous tokens. Right: Draft mechanism remasks invalid tokens within the window and selects future tokens on the right side of the current window exceeding  $\tau_{\text{draft}}$ as placeholders for subsequent decoding.}
    \label{fig:framework}
    \vspace{-0.5cm}
\end{figure}

\begin{figure*}[h!]
    \centering
    \includegraphics[width=0.8\textwidth]{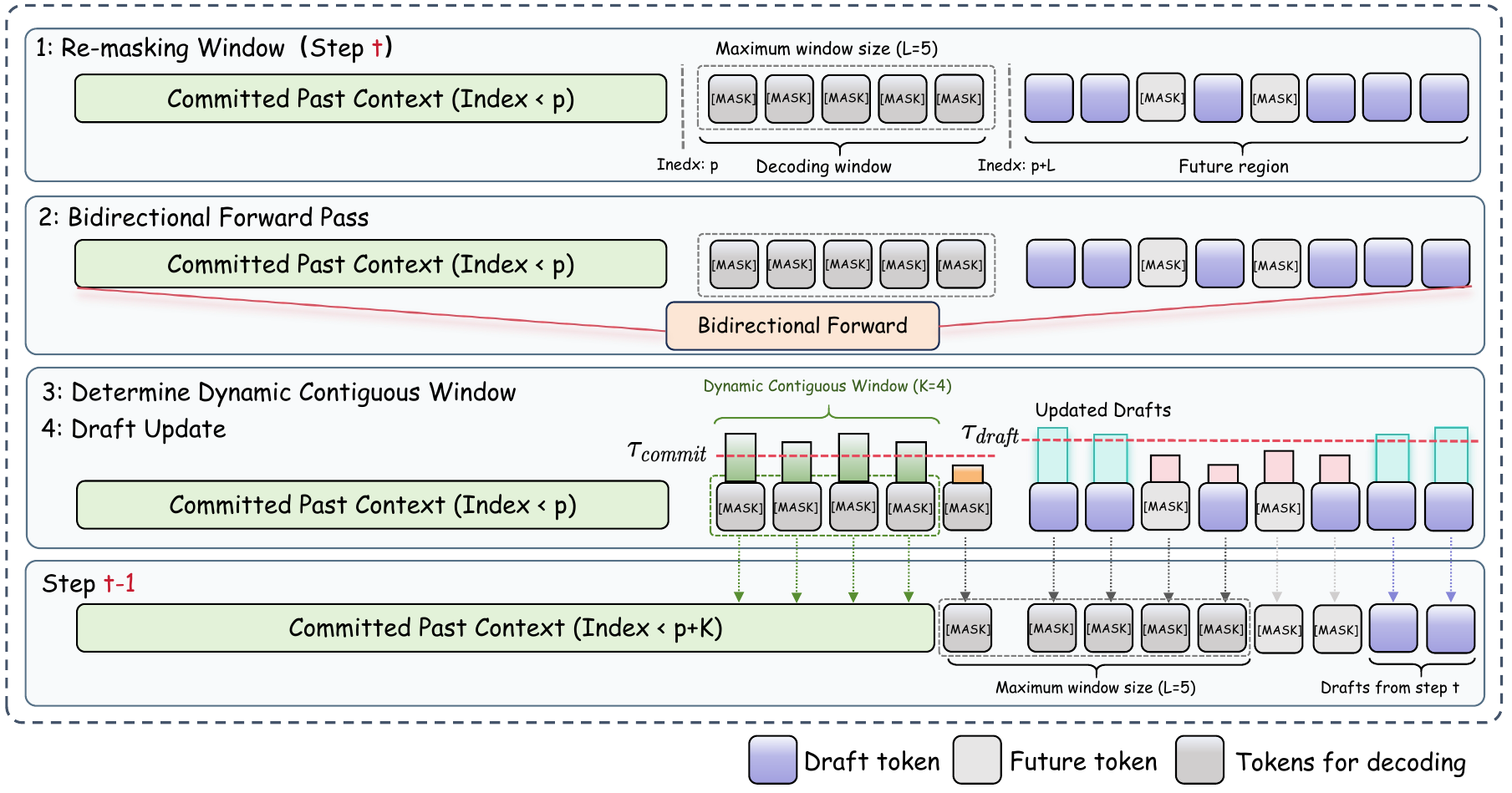}
    \caption{\textbf{Illustration of the draft-guided decoding process.} The proposed pipeline reconciles window-level sequentiality with global contextual awareness. By combining dynamic contiguous verification with an in-place draft update mechanism, DC-Leap utilizes future context on the right side to guide current denoising, effectively alleviating context limitations induced by strictly sequential decoding.}
    \label{fig:flowchart}
\end{figure*}

\subsection{Overview}
Inspired by insights discussed in Sec.~\ref{sec:key_observations}, we propose DC-Leap to safely lower the confidence threshold used in parallel decoding, thus enabling significant inference acceleration for dLLMs. As illustrated in Figure~\ref{fig:framework}, DC-Leap addresses the independence-induced errors common in conventional parallel decoding by incorporating two key components.

First, we introduce \textit{Dynamic Contiguous Verification} (Sec.~\ref{sec:contiguous_verification}). 
This strategy adopts a dynamic decoding window that advances sequentially, while tokens within each window are decoded in parallel. 
By enforcing sequential dependencies at the window level rather than globally, DC-Leap relaxes the restrictive global conditional independence assumption inherent in parallel decoding. As a result, a more lenient confidence threshold can be safely used, allowing more tokens to be accepted at each iteration.

Secondly, we address the issue raised in Sec.~\ref{sec:key_observations} that contiguous decoding may lead bidirectional dLLMs into a suboptimal inference regime by introducing the \textit{draft} mechanism (Sec.~\ref{sec:draft_bank}). Drafts are high-confidence, decoded yet uncommitted tokens outside the current decoding window. These drafts provide valuable look-ahead future context, enabling dLLMs to effectively leverage bidirectional attention capabilities for decoding.

The comprehensive algorithmic procedure of DC-Leap is systematically detailed in Algorithm~\ref{sec:algorithm}.

\subsection{Dynamic Contiguous Verification}
\label{sec:contiguous_verification}
To mitigate Joint Probability Dependence Error (JPDE) without the aggressive confidence filtering used in prior works~\cite{fast-dllm,local-leap}, we propose \textit{Dynamic Contiguous Verification (DCV)}. DCV replaces the global conditional independence assumption inherent to parallel decoding with a local, window-level factorization. By sequentially advancing a contiguous window while decoding all tokens within it in parallel, DCV retains local sequential constraints to mitigate JPDE without sacrificing efficiency. This structural refinement allows the model to adopt a significantly more lenient confidence threshold, effectively capturing acceleration opportunities in lower-confidence regions that were previously suppressed.

\paragraph{Dynamic Decoding Window.} 
We define the dynamic decoding window (DDW) as the set of tokens that will be decoded and committed at the current iteration. 
Unlike fixed-length windows, the span of DDW is dynamically determined from the model’s current predictions, subject to a maximum window size $L$. 

Formally, let $p$ denote the leftmost uncommitted position in the sequence. At the current iteration, the decoding window is a contiguous segment $W = \left [ p, p+K \right ) $, where the window length $K$ satisfies $0 \le K \le L$. 

To dynamically determine the window length $K$ based on the token predictions, we introduce a commitment threshold $\tau_{\text{commit}}$. A token at position $i$ is eligible for commitment if its confidence $c_i \ge \tau_{\text{commit}}$. 
However, unlike standard parallel decoding schemes that commit all tokens exceeding the threshold in a single step, we enforce contiguous commitment: only the longest prefix starting at position $p$ whose tokens all satisfy the confidence constraint is committed.

\begin{table*}[t]
\centering
\footnotesize
\renewcommand{\arraystretch}{0.9}
\caption{\textbf{Performance on Dream-v0-7B-Instruct \cite{dream7b}.} Each cell reports accuracy and throughput (TPS), alongside the speedup factor relative to the baseline. Bold indicates the highest performance.}
\label{tab:dream} 
\resizebox{0.9\linewidth}{!}{
\begin{tabular*}{\linewidth}{@{\extracolsep{\fill}}lcccccc} 
\toprule
& \begin{tabular}[m]{@{}l@{}}MATH  (4-shot)\end{tabular} 
& \begin{tabular}[m]{@{}l@{}}MBPP (3-shot)\end{tabular} 
& \begin{tabular}[m]{@{}l@{}}GSM8K (5-shot)\end{tabular} 
& \begin{tabular}[m]{@{}l@{}}IFEval (0-shot)\end{tabular} 
& \begin{tabular}[m]{@{}l@{}}HumanEval (0-shot) \end{tabular} 
& Avg. \\
\midrule
\textbf{Dream-v0-7B-Instruct} & \textbf{45.44} & 53.40 & \textbf{75.36} & 48.92 & 54.88 & 55.60 \\
\quad Throughput (TPS, $\uparrow$) & 16.87 (1.0$\times$) & 17.37 (1.0$\times$) & 13.92 (1.0$\times$) & 27.25 (1.0$\times$) & 20.17 (1.0$\times$) & 1.0$\times$ \\
\midrule
+ Fast-dLLM & 44.74 & 53.40 & 74.75 & 47.24 & 56.71 & 55.37\\
\quad Throughput (TPS, $\uparrow$) & 46.11 (2.73$\times$) & 46.41 (2.67$\times$) & 44.99 (3.23$\times$) & 38.30 (1.41$\times$) & 37.32 (1.85$\times$) & 2.38$\times$ \\
\midrule
+ LocalLeap & 44.66 & 57.00 & 73.92 & 50.00 & 56.71 & 56.46\\
\quad Throughput (TPS, $\uparrow$) & 51.44 (3.05$\times$) & 81.75 (4.71$\times$) & 57.68 (4.14$\times$) & 77.37 (2.84$\times$) & 83.26 (4.13$\times$) & 3.77$\times$ \\
\midrule
+ L2P & 44.98 & 51.60 & 73.92 & 45.76 & 58.54 & 54.96\\
\quad Throughput (TPS, $\uparrow$) & 20.37 (1.21$\times$) & 30.19 (1.74$\times$) & 23.65 (1.70$\times$) & 48.62 (1.78$\times$) & 40.90 (2.03$\times$) & 1.69$\times$ \\
\midrule
+ DC-Leap (Ours) & 44.18 & \textbf{57.60} & 73.92 & \textbf{51.92} & \textbf{59.76} & \textbf{57.48}\\
\quad Throughput (TPS, $\uparrow$) & \textbf{53.70} (\textbf{3.18$\times$}) & \textbf{88.59} (\textbf{5.10$\times$}) & \textbf{75.87} (\textbf{5.45$\times$}) & \textbf{110.64} (\textbf{4.06$\times$}) & \textbf{116.21} (\textbf{5.76$\times$}) & \textbf{4.71}$\times$\\
\bottomrule
\end{tabular*}
}
\end{table*}

Consequently, within the maximum window size $L$, the window length $K$ is determined by truncating at the first position where the confidence falls below $\tau_{\text{commit}}$.
For implementation convenience, we compute $K$ using an indicator-based prefix-product formulation:
\begin{equation}
\label{eqn:compute_k}
    K = \sum_{i=0}^{L-1} \prod_{j=0}^{i} \mathbb{I}(c_{j} > \tau_{\text{commit}}),
\end{equation}
which counts the length of the longest contiguous prefix whose token confidences exceed the commit threshold.

This sequential advancement of the contiguous and monotonic window ensures that tokens are verified in their correct order. By preventing out-of-order commitments, the mechanism allows DCV to safely adopt a more lenient confidence threshold, thereby maximizing the number of tokens accepted in each parallel decoding iteration.

\subsection{Draft-guided Decoding}
\label{sec:draft_bank}
While DCV effectively mitigates JPDE through window-level sequentiality, it inherently restricts the model's access to right-side context during inference. Such context truncation diminishes the ability of bidirectional dLLMs to exploit future token information, potentially leading to suboptimal inference performance. To overcome this issue, we introduce \textit{Draft-guided Decoding}, a method that maintains uncommitted draft placeholders to provide look-ahead context, thus preserving the structural advantages of bidirectional attention during inference.

Specifically, we generate \textit{drafts} by predicting tokens at select positions beyond the current decoding window, employing them as auxiliary right-side context. These drafts are maintained as \textit{look-ahead placeholders} and are not finalized as outputs, thus preserving window-level sequentiality. In essence, drafts serve as right-side contextual anchors for decoding the current window, aligning inference-time conditions more closely with the bidirectional denoising setting encountered during training, and thereby enhancing the effective utilization of bidirectional attention.

\begin{table*}[t]
\centering
\footnotesize
\renewcommand{\arraystretch}{0.9}
\caption{\textbf{Performance on LLaDA-8B-Instruct \cite{llada}.} Each cell reports accuracy and throughput (TPS), alongside the speedup factor relative to the baseline. Bold indicates the highest performance.}
\label{tab:llada}  
\resizebox{0.9\linewidth}{!}{
\begin{tabular*}{\linewidth}{@{\extracolsep{\fill}}lcccccc} 
\toprule
& \begin{tabular}[m]{@{}l@{}}MATH (4-shot)\end{tabular} 
& \begin{tabular}[m]{@{}l@{}}MBPP (3-shot)\end{tabular} 
& \begin{tabular}[m]{@{}l@{}}GSM8K (5-shot)\end{tabular} 
& \begin{tabular}[m]{@{}l@{}}IFEval (0-shot)\end{tabular} 
& \begin{tabular}[m]{@{}l@{}}HumanEval (0-shot)\end{tabular} 
& Avg. \\
\midrule
\textbf{LLaDA-8B-Instruct} & \textbf{38.74} & 30.00 & 77.71 & 70.74 & \textbf{35.37} & 50.51 \\
\quad Throughput (TPS, $\uparrow$) & 3.40 (1.0$\times$) & 12.96 (1.0$\times$) & 10.47 (1.0$\times$) & 21.07 (1.0$\times$) & 19.16 (1.0$\times$) & 1.0$\times$ \\
\midrule
+ Fast-dLLM & 38.30 & 30.00 & \textbf{77.94} & 70.62 & \textbf{35.37} & \textbf{50.45}\\
\quad Throughput (TPS, $\uparrow$) & 33.08 (9.73$\times$) & 44.93 (3.47$\times$) & 31.90 (3.05$\times$) & 49.14 (2.33$\times$) & 57.65 (3.01$\times$) & 4.32$\times$ \\
\midrule
+ LocalLeap & 37.94 & 30.00 & 77.33 & 70.14 & 34.76 & 50.03\\
\quad Throughput (TPS, $\uparrow$) & 41.97 (12.34$\times$) & 55.98 (4.32$\times$) & 41.86 (4.00$\times$) & 54.88 (2.60$\times$) & 76.33 (3.98$\times$) & 5.45$\times$ \\
\midrule
+ L2P & 37.96 & \textbf{30.60} & 77.41 & 70.26 & 34.76 & 50.20\\
\quad Throughput (TPS, $\uparrow$) & 44.35 (13.04$\times$) & 57.77 (4.46$\times$) & 43.19 (4.13$\times$) & 59.13 (2.81$\times$) & 80.57 (4.21$\times$) & 5.73$\times$ \\
\midrule
+ DC-Leap (Ours) & 37.28 & 29.60 & 76.65 & \textbf{71.34} & \textbf{35.37} & 50.05\\
\quad Throughput (TPS, $\uparrow$) & \textbf{49.82} (\textbf{14.65$\times$}) & \textbf{59.30} (\textbf{4.58}$\times$) & \textbf{50.22} (\textbf{4.80$\times$}) & \textbf{72.48} (\textbf{3.44}$\times$) & \textbf{90.72} (\textbf{4.73$\times$}) & \textbf{6.44}$\times$\\
\bottomrule
\end{tabular*}
}
\end{table*}
\begin{table*}[t]
\centering
\footnotesize
\caption{\textbf{Performance on LLaDA-1.5 \cite{llada1.5}.} Each cell reports accuracy and throughput (TPS), alongside the speedup factor relative to the baseline. Bold indicates the highest performance.}
\label{tab:llada1.5}  
\resizebox{0.9\linewidth}{!}{
\renewcommand{\arraystretch}{0.85}
\begin{tabular*}{\linewidth}{@{\extracolsep{\fill}}lcccccc} 
\toprule
& \begin{tabular}[m]{@{}l@{}}MATH (4-shot)\end{tabular} 
& \begin{tabular}[m]{@{}l@{}}MBPP (3-shot)\end{tabular} 
& \begin{tabular}[m]{@{}l@{}}GSM8K (5-shot)\end{tabular} 
& \begin{tabular}[m]{@{}l@{}}IFEval (0-shot)\end{tabular} 
& \begin{tabular}[m]{@{}l@{}}HumanEval (0-shot)\end{tabular} 
& Avg. \\
\midrule
\textbf{LLaDA-1.5} & \textbf{38.96} & 37.40 & 80.74 & 66.67 & \textbf{40.24} & 52.80 \\
\quad Throughput (TPS, $\uparrow$) & 3.46 (1.0$\times$) & 13.14 (1.0$\times$) & 10.58 (1.0$\times$) & 21.65 (1.0$\times$) & 19.38 (1.0$\times$) & 1.0$\times$ \\
\midrule
+ Fast-dLLM & 38.94 & 37.60 & 81.05 & 66.97 & 39.63 & 52.84\\
\quad Throughput (TPS, $\uparrow$) & 33.80 (9.77$\times$) & 73.83 (5.62$\times$) & 32.92 (3.11$\times$) & 48.39 (2.24$\times$) & 55.34 (2.86$\times$) & 4.72$\times$ \\
\midrule
+ LocalLeap & 38.22 & 39.20 & \textbf{81.58} & 65.11 & \textbf{40.24} & 52.87\\
\quad Throughput (TPS, $\uparrow$) & 42.28 (12.22$\times$) & 90.06 (6.85$\times$) & 43.51 (4.11$\times$) & 68.55 (3.17$\times$) & 71.28 (3.68$\times$) & 6.01$\times$ \\
\midrule
+ L2P & 38.66 & 37.40 & 80.89 & 66.31 & 39.02 & 52.46\\
\quad Throughput (TPS, $\uparrow$) & 45.30 (13.09$\times$) & 96.43 (7.34$\times$) & 45.68 (4.32$\times$) & 56.83 (2.62$\times$) & 76.00 (3.92$\times$) & 6.26$\times$ \\
\midrule
+ DC-Leap (Ours) & 38.66 & \textbf{42.60} & 80.21 & \textbf{71.82} & 38.41 & \textbf{54.34}\\
\quad Throughput (TPS, $\uparrow$) & \textbf{51.72} (\textbf{14.95}$\times$) & \textbf{171.35} (\textbf{13.04}$\times$) & \textbf{52.07} (\textbf{4.92}$\times$) & \textbf{69.67} (\textbf{3.22}$\times$) & \textbf{102.74} (\textbf{5.30}$\times$) & \textbf{8.29}$\times$\\
\bottomrule
\end{tabular*}
}
\end{table*}

\paragraph{Draft Update.}
We now formalize the draft update procedure. During the forward pass, predictions in the future region are monitored in parallel. We introduce a draft threshold $\tau_{\text{draft}}$ and designate positions whose confidence satisfies $c > \tau_{\text{draft}}$ as draft candidates. 
After committing the tokens within the decoding window, we update the sequence by overwriting the corresponding draft placeholders in place. 

Formally, let $\hat{x}_i$ denote the top-1 predicted token at position $i$ in the future region, \emph{i.e.}, $i \ge p + L$, where $p$ is the index of the last verified token and $L$ is the decoding window length. Let $c_i$ denote the confidence score of token $\hat{x}_i$. We maintain a draft placeholder $d_i$ for each future position $i$. Initially set as a \texttt{[MASK]} token, $d_i$ is iteratively updated as follows: 
\begin{equation}
\label{eqn:draft_update}
    d_i = 
    \begin{cases} 
    \hat{x}_i & \text{if } c_i > \tau_{\text{draft}} \\
    d_i & \text{otherwise}
    \end{cases}
    \quad, \forall i \ge p+L,
\end{equation}

\paragraph{Draft Invalidation.}
However, since drafts are non-committed placeholders, they must be invalidated once they fall within the range of tokens eligible for inclusion in the decoding window. To prevent trivial copying and to preserve window-level sequential decoding, we discard such drafts by re-masking. For simplicity, we re-mask all tokens  within this maximum length range, thereby removing any drafts that are potentially incorporated into the decoding window:
\begin{equation}
\label{eqn:remasking}
    x_{p+i} \Leftarrow  \texttt{[MASK]}, \quad \forall i \in [0, L-1].
\end{equation}
Here, $p$ denotes the starting position of the current decoding window, $L$ denotes the maximum window size, and $x_{p+i}$ is the token at position $p+i$ among those eligible for inclusion in the decoding window.

\paragraph{Inference with Drafts.} 

At each decoding iteration, the proposed draft-guided decoding pipeline consists of four steps as shown in Figure~\ref{fig:flowchart}: (i) re-masking the decoding window to invalidate any drafts it contains as Eq.~\eqref{eqn:remasking}; (ii) performing a single forward pass to obtain predictions for both the window and the future region; (iii) computing the dynamic contiguous window length $K$ as Eq.~\eqref{eqn:compute_k}; and (iv) committing the verified window tokens while updating drafts in place as Eq.~\eqref{eqn:draft_update}. 
Overall, the draft-guided decoding pipeline provides look-ahead future context to the decoding region, while still preserving the window-level sequentiality of DCV, thereby alleviating the distributional mismatch and context limitations induced by the sequential decoding.

\section{Experiments}
\subsection{Experiments Setup}

\begin{figure*}[]
    \centering
    \begin{subfigure}[b]{0.24\textwidth}
        \centering
        \includegraphics[width=\textwidth]{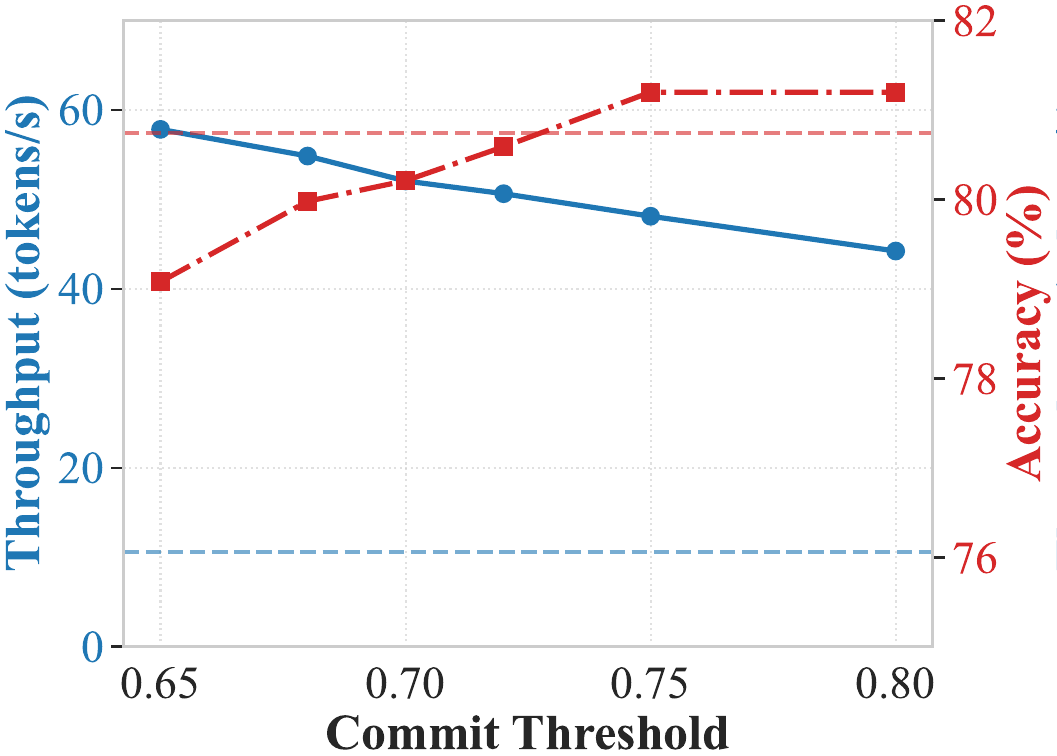} 
        \caption{GSM8K: Sensitivity to Commitment Threshold}
        \label{fig:gsm8k_commit}
    \end{subfigure}
    \hfill
    \begin{subfigure}[b]{0.24\textwidth}
        \centering
        \includegraphics[width=\textwidth]{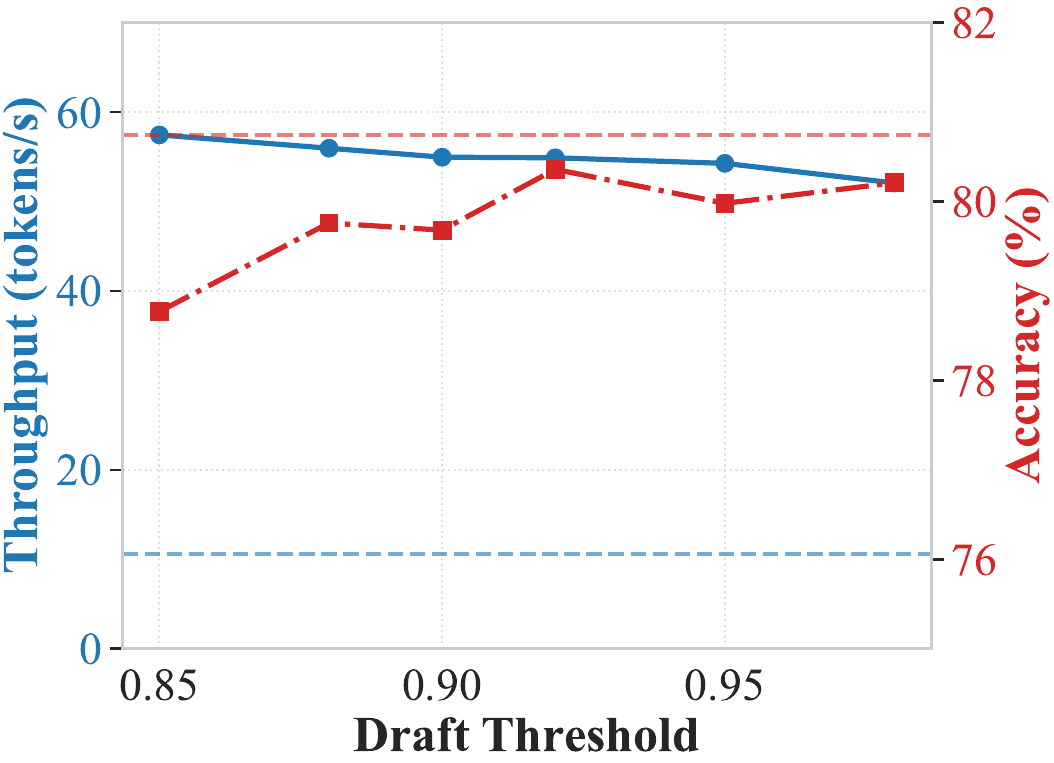}
        \caption{GSM8K: Sensitivity to Draft Threshold}
        \label{fig:gsm8k_draft}
    \end{subfigure}
    \hfill
    \begin{subfigure}[b]{0.24\textwidth}
        \centering
        \includegraphics[width=\textwidth]{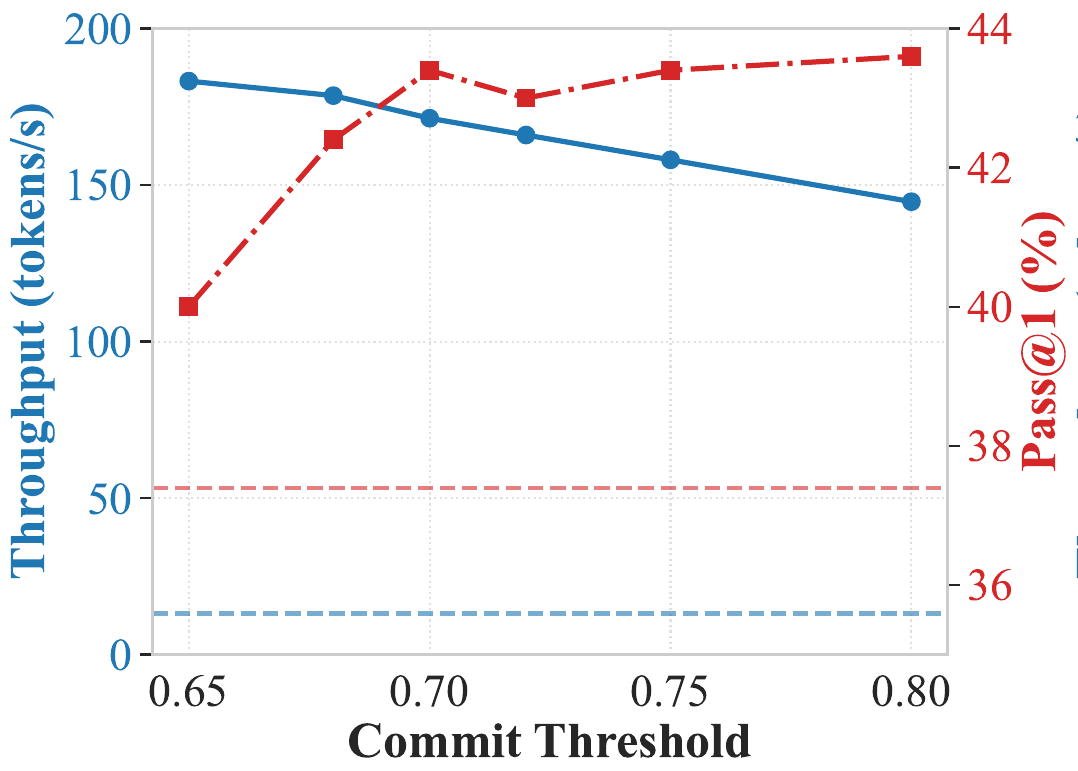}
        \caption{MBPP: Sensitivity to Commitment Threshold}
        \label{fig:mbpp_commit}
    \end{subfigure}
    \hfill
    \begin{subfigure}[b]{0.24\textwidth}
        \centering
        \includegraphics[width=\textwidth]{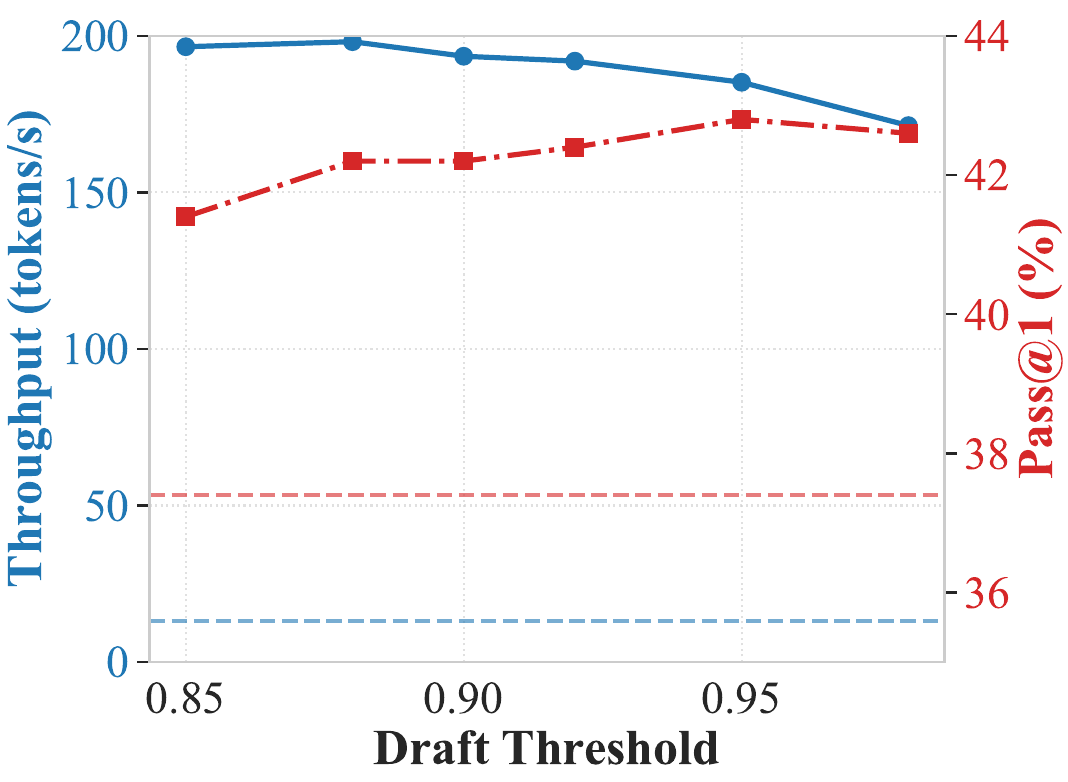}
        \caption{MBPP: Sensitivity to Draft Threshold}
        \label{fig:mbpp_draft}
    \end{subfigure}
    
   \caption{\textbf{Sensitivity analysis on $\tau_{\text{c}}$ and  $\tau_{\text{d}}$ for LLaDA-1.5 on GSM8K and MBPP benchmarks.}
   We perform univariate analysis for each threshold while fixing the other at the default setting. Blue solid lines and red dash-dotted lines represent throughput (TPS) and task performance (Accuracy or Pass@1), respectively, with horizontal dashed lines marking corresponding baseline levels.}
    \label{fig:ablation_sensitivity}
\end{figure*}

\begin{figure}[t!]
    \centering
    \includegraphics[width=0.8\linewidth]{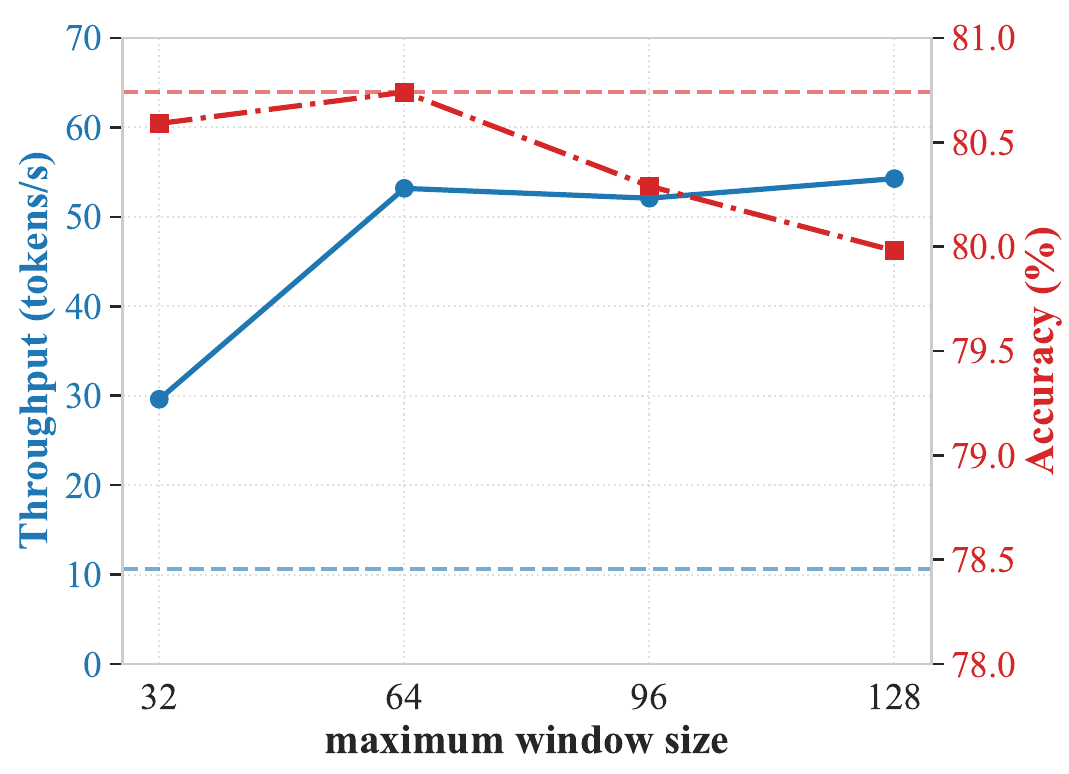}
    \caption{\textbf{Sensitivity analysis on the maximun window lehgth $L$ for LLaDA-1.5 on GSM8K benchmark.} Throughput (TPS, blue solid) and accuracy (red dash-dotted) are plotted against $L$, with horizontal dashed lines marking their respective baselines.}
    \label{fig:window_size}
\end{figure}

\paragraph{Models and Baselines.}
We select LLaDA-8B-Instruct ~\cite{llada}, LLaDA-1.5~\cite{llada1.5}, and Dream-v0-Instruct-7B~\cite{dream7b} as our base models. For each, the generation follows a semi-autoregressive remasking strategy~\cite{llada} configured with a block size of 32. We compare our DC-Leap against standard diffusion decoding and three state-of-the-art parallel decoding strategies: 
(1) Fast-dLLM~\cite{fast-dllm}, a confidence-thresholding parallel decoding method; (2) LocalLeap~\cite{local-leap}, which leverages high-confidence anchors to guide local parallelization; and 
(3) L2P~\cite{l2p}, a learnable policy network designed for adaptive parallel decoding.

\paragraph{Benchmarks and Metrics.} 
To comprehensively assess the performance and efficiency of our method, we conduct evaluations across three primary domains: Mathematical Reasoning (GSM8K~\cite{gsm8k}, MATH~\cite{math}), Program Synthesis (HumanEval~\cite{humaneval}, 
MBPP~\cite{mbpp}), and Instruction Following (IFEval~\cite{ifeval}). We set the generation length to 512 for HumanEval and IFEval, while a length of 256 is applied to all other benchmarks. Efficiency is quantified via Tokens Per Second 
(TPS) and relative speedup over the baseline.

\paragraph{Implementation Details.} All experiments are performed on NVIDIA RTX PRO 6000 Blackwell. To ensure reproducibility and fair comparison, we utilize greedy decoding without any random sampling strategies. Unless otherwise specified, 
we set the maximum decoding window size to half of the total generation length, which determines the span of tokens being verified. Throughout our experiments, we maintain constant threshold values of $\tau_{\text{commit}}$ = 0.70 and $\tau_{\text{draft}}$ = 0.98 across all models and benchmarks, demonstrating the robustness and generalization of DC-Leap in diverse evaluation scenarios.

\subsection{Main Results}
The results are presented in Table~\ref{tab:dream},~\ref{tab:llada}, and~\ref{tab:llada1.5}. Overall, DC-Leap achieves substantial inference speedup with negligible performance degradation and, in several cases, slight improvement. In terms of decoding speed, DC-Leap consistently outperforms all baselines across diverse models and benchmarks. On Dream-v0-7B-Instruct (Table~\ref{tab:dream}), DC-Leap achieves an average speedup of $\mathbf{4.71\times}$, significantly surpassing the previous state-of-the-art LocalLeap ($3.77\times$) and Fast-dLLM ($2.38\times$). The performance gain is even more noticeable on the LLaDA-1.5 (Table~\ref{tab:llada1.5}). By unlocking the potential of lower-confidence tokens via contiguous verification, DC-Leap attains an average speedup of $\mathbf{8.29\times}$. Notably, on code generation tasks such as MBPP, DC-Leap achieves a throughput improvement of $\mathbf{13.04\times}$, demonstrating its exceptional efficiency.

\paragraph{Scalability and Extension.}

We further evaluate DC-Leap on long sequence generation (1024 tokens) and its compatibility with KV cache on LLaDA-1.5, as shown in Table~\ref{tab:llada1.5_1024_cache}. The results highlight superior scalability, where acceleration gains amplify with sequence length. 
Specifically, on GSM8K and MBPP, DC-Leap achieves \textbf{24.64$\times$} and \textbf{53.18 $\times$} speedup,  respectively. Furthermore, DC-Leap proves orthogonal to the KV Cache optimizations. Integrating DC-Leap with \textbf{dLLM-Cache}~\cite{dllm-cache} 
yields a compound acceleration effect, pushing throughput to a \textbf{105.02 $\times$ (809.71 TPS)} on MBPP while maintaining competitive generation quality. This establishes DC-Leap as a robust, plug-and-play module for dLLMs acceleration.

\subsection{Ablations and Analysis}

\label{sec:ablation}

To understand hyperparameter sensitivity in DC-Leap, we conduct extensive ablations on LLaDA-1.5 using GSM8K and MBPP, analyzing the impact of dual thresholds (commitment and draft) and maximum window size.

\paragraph{Impact of commitment threshold $\tau_c$.} 
To explore the impact of dynamic contiguous verification, we fix the draft threshold $ \tau_d $ at 0.98 and vary $\tau_c$. As shown in Figure~\ref{fig:ablation_sensitivity} (a, c), lowering $\tau_c$ unlocks significant speedups by allowing more tokens to be accepted within the decoding window. Specifically, on GSM8K, decreasing $\tau_c$ from $0.80$ to $0.65$ boosts throughput from $44.24$ to $57.84$ TPS without compromising accuracy, demonstrating that DC-Leap is robust to $\tau_c$. A consistent trend is observed in the MBPP results. To strike an optimal balance between accuracy and throughput, we set $\tau_c = 0.70$ as our default configuration.
\paragraph{Impact of draft threshold $\tau_d$.} 
Similarly, we analyze the impact of $\tau_d$ on accuracy and throughput in Figures~\ref{fig:ablation_sensitivity} (b, d), while fixing $\tau_c$ to 0.70. The performance on both benchmarks follows a consistent pattern: Lowering $\tau_d$ compromises accuracy slightly. Conversely, increasing $\tau_d$ slightly reduces throughput due to stricter draft selection, as seen in MBPP, where TPS adjusts from 185.19 to 171.35. To ensure throughput and consistent performance across tasks, we identify $\tau_d = 0.98$ as the optimal value.

\paragraph{Impact of maximum window size $L$.}
We investigate the sensitivity of the maximum window size $L$ on the GSM8K benchmark, varying from 32 to 128. As illustrated in Figure~\ref{fig:window_size}, expanding the window size significantly enhances parallelism. While further increasing to 128 maximizes inference speed, reaching a peak throughput of 54.27 TPS, at the cost of a slight accuracy decline from 80.74\% to 79.98\%. Given that the performance degradation is negligible relative to the substantial efficiency gains, we adopt $L$ = 128 as the default setting to prioritize acceleration.

\begin{table}[t!]
\centering
\caption{\textbf{Performance of DC-Leap on 1024-token generation, with and without dLLM-Cache.} Best performances are in bold.}
\label{tab:llada1.5_1024_cache}
\renewcommand{\arraystretch}{1.0} 
\resizebox{\columnwidth}{!}{
    \begin{tabular}{llcc}
    \toprule
     & \textbf{Method} 
     & \textbf{Throughput (TPS, $\uparrow$)} 
     & \textbf{Performance (Score \%)} \\
    \midrule
    \textbf{GSM8K} & LLaDA-1.5 & 6.63 (1.0$\times$) & 82.79 \\
                   & + Cache   & 17.23 (2.60$\times$) &  \textbf{83.09}\\
                   & + DC-Leap & 163.38 (24.64$\times$) & 82.34 \\
                   & + DC-Leap + Cache & \textbf{404.38 (60.99$\times$)} & 79.30 \\ 
    \midrule
    \textbf{MBPP}  & LLaDA-1.5 & 7.71 (1.0$\times$) & 38.40 \\
                   & + Cache   & 18.64 (2.42$\times$) & \textbf{39.40} \\ 
                   & + DC-Leap & 410.07 (53.19$\times$) & \textbf{39.40} \\
                   & + DC-Leap + Cache & \textbf{809.71 (105.02$\times$)} & 36.40 \\ 
    \bottomrule
    \end{tabular}
}
\end{table}
\section{Conclusion}

In this work, we identify the strict thresholding driven by the conditional independence assumption in non-contiguous decoding as a critical bottleneck limiting inference speed. While sequential decoding theoretically resolves this issue, it remains inefficient and underutilizes bidirectional context. Building on these insights, we propose DC-Leap, a simple yet effective method that introduces Dynamic Contiguous Verification and leverages the inherent bidirectional attention of dLLMs via Draft-guided Decoding to unlock the acceleration potential within lower-confidence regimes. As a plug-and-play method, DC-Leap is orthogonal to existing KV Cache optimizations, allowing their acceleration effects to be stacked for synergistic gains.

\section*{Impact Statement}
This paper presents work whose goal is to advance the field of machine learning. There are many potential societal consequences of our work, none of which we feel must be specifically highlighted here.

\section*{Acknowledgement}
This work was supported by the Guangdong Basic and Applied Basic Research Foundation (2025A1515011546) and by the Shenzhen Science and Technology Program (JCYJ20240813105901003, ZDCY20250901113000001).
\bibliographystyle{icml2026}
\bibliography{references}

\newpage
\appendix
\onecolumn

\AppTOCTitle
\vspace{-0.5em}

\apptocA{sec:algorithm}{A \quad Algorithm of DC-Leap}

\apptocA{sec:app_b}{B \quad Related Work}
\apptocB{sec:app_b1}{B.1 \quad Inference Acceleration in LLMs and VLMs}
\apptocB{sec:app_b2}{B.2 \quad Diffusion Large Language Models}
\apptocB{sec:app_b3}{B.3 \quad Sampling and Acceleration Strategies of dLLMs}

\apptocA{sec:app_c}{C \quad Case Study}

\apptocA{sec:app_d}{D \quad Analysis of Decoding Dynamics}

\apptocA{sec:app_e}{E \quad Analysis of Continuous Drafts}
\apptocB{sec:app_e1}{E.1 \quad Motivation and Hypothesis}
\apptocB{sec:app_e2}{E.2 \quad Experimental Setup}
\apptocB{sec:app_e3}{E.3 \quad Results}
\apptocB{sec:app_e4}{E.4 \quad Observations and Discussions}

\apptocA{sec:app_f}{F \quad More Results and Analysis}
\apptocB{sec:app_f1}{F.1 \quad More Results of Ablations}
\apptocB{sec:app_f2}{F.2 \quad More Results of Step Distilled dLLMs}
\apptocB{sec:app_f3}{F.3 \quad Analysis of Alternative Confidence Definition}

\vspace{1cm}
\hrule
\vspace{1cm}

\newpage
\section{Algorithm of DC-Leap}
\label{sec:algorithm}
The pseudocode of DC-Leap is shown in Algorithm~\ref{alg:dc_leap}.
\begin{algorithm}[H]
    \caption{DC-Leap: Draft-Guided Contiguous Parallel Decoding}
    \label{alg:dc_leap}
    \begin{algorithmic}[1]
        \REQUIRE Model $p_\theta$, prompt $\mathbf{x}_{\text{prompt}}$, generation length $L_\text{g}$, max window size $L$,confidence  thresholds $\tau_{\text{commit}}, \tau_{\text{draft}}$
        \STATE Initialize sequence $\mathbf{x} \leftarrow \text{concat}(\mathbf{x}_{\text{prompt}}, \{\texttt{[MASK]}\}^{L_\text{g}})$
        \STATE $p \leftarrow \text{len}(\mathbf{x}_{\text{prompt}})$ \hfill $\triangleright$ Pointer to the first unverified position
        
        \WHILE{$p < L_g$}
            \STATE \textbf{// Draft Invalidation}
            \STATE $\mathbf{x}_{p:p+L} \leftarrow \{\texttt{[MASK]}\}^L$ \hfill 
            
            \STATE \textbf{// Forward Pass}
            \STATE $\hat{\mathbf{x}}, \mathbf{c} \leftarrow \text{Forward}(p_\theta, \mathbf{x})$ \hfill
            
            \STATE \textbf{// Dynamic Contiguous Verification}
            \STATE $K \leftarrow \sum_{i=0}^{L-1} \prod_{j=0}^{i} \mathbb{I}(c_{p+j} > \tau_{\text{commit}})$ \hfill $\triangleright$  Caculate the length of dynamic decoding window
            
            \STATE \textbf{// Commit Verified Tokens \& Update Draft Placeholders}
            \STATE $\mathbf{x}_{p:p+K} \leftarrow \hat{\mathbf{x}}_{p:p+K}$ \hfill $\triangleright$ Commit verified tokens
            \FOR{$j = p+L$ \textbf{to} $L_g$}
                \IF{$c_j > \tau_{\text{draft}}$}
                    \STATE $x_j \leftarrow \hat{x}_j$ \hfill $\triangleright$  Update future drafts
                \ENDIF
            \ENDFOR
            
            \STATE $p \leftarrow p + K$ \hfill $\triangleright$ Dynamically advance
        \ENDWHILE
        \STATE \textbf{return} $\mathbf{x}$
    \end{algorithmic}
\end{algorithm}
\section{Related Work}
\label{sec:app_b}
\subsection{Inference Acceleration in LLMs and VLMs}
\label{sec:app_b1}

Inference efficiency remains a critical bottleneck for both Large Language Models (LLMs) and Vision Language Models (VLMs). 

In autoregressive (AR) LLMs such as Llama 3~\cite{llama3}, Qwen 3~\cite{qwen3} and Mistral 7B~\cite{mistral7b}, acceleration strategies primarily target the sequential bottleneck of the decoding process. While advancements in preference optimization and continual learning bolster complex reasoning capabilities~\cite{lai2024step, peng2024scalable, peng2025omni}, inference acceleration has also become increasingly vital. Speculative decoding emerges as a representative paradigm, where a lightweight drafter predicts multiple future tokens that are subsequently verified in parallel by a larger target model~\cite{speculative_decoding, speculative_sampling}. Building on this concept, recent works have introduced multi-head structures and tree-based verification to further expand the parallel search space during generation~\cite{medusa, eagle}. Additionally, quantization and pruning techniques reduce the memory footprint and computational intensity, allowing models to achieve higher throughput on resource-constrained hardware~\cite{gptq, awq}.

For VLMs, such as Qwen3-VL~\cite{qwen3vl}, LLaVa-Next~\cite{llavanext} and InternVL 2.5~\cite{internvl2.5}, inference acceleration is further complicated by the massive number of visual tokens processed through cross-modal attention. Current research addresses this by identifying and removing redundant visual information via token pruning or merging strategies~\cite{tome, NAT, yang2025visionzip}. Dynamic resolution techniques also allow models to allocate more computational resources to informative image regions while processing background areas at a lower cost~\cite{dvit}. Furthermore, multimodal speculative decoding has been proposed to exploit the correlation between visual features and linguistic outputs, enabling the parallel drafting of vision-textual sequences~\cite{v_spec}. Such efficiencies are vital for advancing large multimodal reasoning models~\cite{li2025perception}, which have recently expanded to complex tasks like reasoning segmentation~\cite{lai2024lisa, yang2023lisa++} and object hallucination mitigation~\cite{peng2025mitigating}. While these methods effectively mitigate the overhead of AR architectures, they remain bound by the causal requirement of sequential drafting. In contrast, dLLMs provide a framework for fully parallel decoding, which can theoretically achieve superior speedups.

Moreover, the principles of efficient contextual processing and robust representation learning are critical across broader computer vision domains. This includes optimizing feature extraction for few-shot and semi-supervised semantic segmentation~\cite{tian2020prior, lai2021semi, tian2022generalized, peng2023hierarchical, luo2023pfenet++}, refining adaptive and context-aware classifiers for long-tailed scenarios~\cite{tian2022adaptive, tian2023learning, cui2022reslt}, and advancing training-free or open-vocabulary dense perception alongside cross-domain adaptation and generalized contrastive learning~\cite{shao2024explore, yang2024unified, wang2025declip, wang2025generalized, cui2023generalized}. Early efforts have also explored efficient shape-aware embeddings for specific downstream tasks like text detection~\cite{tian2019learning}. Similarly, in 3D understanding, parallel challenges in computational overhead have spurred the development of sparse and adaptive architectures for point cloud segmentation~\cite{jiang2021guided, ning2023boosting, peng2024oa}, alongside scalable multi-dataset prompt training~\cite{wu2024ppt} and joint 2D-3D self-supervised representation frameworks~\cite{wang2024groupcontrast, zhang2025concerto}. Efficient representations in these spaces are also key to downstream generation tasks, such as multi-angle 3D asset editing~\cite{huang2025edit360}.

\subsection{Diffusion Large Language Models}
\label{sec:app_b2}
The paradigm of generative diffusion models, originally dominant in continuous domains such as image synthesis, has recently been adapted to the discrete nature of natural language. Early attempts, such as Diffusion-LM~\cite{diffusionLM}, mapped discrete tokens 
to a continuous embedding space to apply Gaussian diffusion, though this often resulted in suboptimal text generation quality due to rounding errors. SSD-LM~\cite{ssdLM} improved upon this by performing diffusion on the probability simplex. However, the field 
has largely converged towards Masked Diffusion Models (MDMs), which operate directly in the discrete state space by treating the forward process as a random masking operation.

Recent scaling efforts have demonstrated that dLLMs can rival the performance of autoregressive (AR) baselines. LLaDA~\cite{llada, llada1.5} and Dream~\cite{dream7b} represent the state-of-the-art in this domain, scaling masked diffusion to billions of parameters. 
Unlike AR models which are constrained by causal attention, these models utilize bidirectional attention, allowing every token to attend to the entire sequence context during inference. While this architecture theoretically supports parallel decoding, the standard sampling process, requiring tens or hundreds of iterative refinement steps, remains a significant bottleneck for latency-sensitive applications.

\subsection{Sampling and Acceleration Strategies of dLLMs}
\label{sec:app_b3}
For dLLMs, acceleration strategies must navigate a different trade-off between parallelism and consistency. Standard samplers like Fast-dLLM~\cite{fast-dllm} introduce training-free acceleration by implementing a confidence-aware parallel decoding strategy, where 
tokens exceeding a certain probability threshold are accepted simultaneously. Building on this, LocalLeap~\cite{local-leap} observes a locality property in decoding and employs high-confidence tokens as anchors to dynamically relax thresholds for neighboring tokens. 
DPad~\cite{dpad} further optimizes efficiency by applying suffix dropout and sliding windows to prune redundant computations in the suffix region.

Concurrently, several recent works have explored more sophisticated adaptive policies for accelerating dLLM inference. UNCODE~\citep{pc-sampler} addresses decoding biases by calibrating confidence scores with positional priors and informativeness priors to mitigate rigid boundary bias and trivial token bias. Prophet~\cite{prophet} introduces an early-commit mechanism that uses the confidence gap between top-two candidates as a criterion to dynamically terminate denosing steps once predictions stabilize. Moving beyond fixed heuristics, Learn2PD~\cite{l2p} proposes a post-training framework that employs a lightweight MLP filter to adaptively predict token stability and optimize remasking decisions.

However, a critical limitation in these parallel approaches is the independence assumption. As noted by the EB-Sampler~\cite{EB-Sampler}, independently unmasking multiple tokens introduces a joint probability dependence error, as the validity of one token often 
depends on the specific instantiation of its neighbors. To mitigate this, prior methods like Fast-dLLM and LocalLeap are forced to adopt conservative, high-confidence thresholds (e.g., $> 0.9$ ) to ensure safety, thereby sacrificing the potential speedup from correctly 
predicted but lower-confidence tokens. Our work addresses this limitation by enforcing a causal, left-to-right verification order, which eliminates the joint dependence error and unlocks the stable utilization of lower confidence predictions.
\section{Case Study}
\label{sec:app_c}

We evaluate the practical impact of DC-Leap through a case study on LLaDA-8B-Instruct, LLaDA-1.5, and Dream-v0-7B-Instruct. Table~\ref{tab:case_study} presents the results both with and without DC-Leap integration.
All experiments are conducted on NVIDIA RTX PRO 6000 Blackwell, with a generation length of 256, a block length of 32, and a maximum decoding window size set to half of the total generation length. 
Using a commitment threshold $\tau_c=0.65$ and a draft threshold $\tau_d=0.95$, empirical results show that DC-Leap achieves substantial wall-clock speedups while maintaining reasoning accuracy across all models.

Interestingly, while the final answers remain identical to the baselines, DC-Leap often yields more concise reasoning paths. This suggests that our draft-guided mechanism allows the model to incorporate future high-confidence tokens through bidirectional attention, providing a global context that guides the model to bypass redundant steps and steer the denoising trajectory towards a more efficient logical sequence.
Furthermore, the absence of hallucinations even with the lower commitment threshold ($\tau_c = 0.65$) demonstrates that our contiguous constraint effectively stabilizes the generation process, ensuring that the leaps in decoding remain logically grounded and instruction-compliant.

\begingroup
\footnotesize  
\setlength{\tabcolsep}{4pt} 
\renewcommand{\arraystretch}{1.1} 

\begin{longtable}{p{0.15\textwidth} p{0.80\textwidth}}
\caption{\textbf{Case Studies across LLaDA-8B-Instruct, LLaDA-1.5, and Dream-v0-7B-Instruct.}} 
\label{tab:case_study} \\
\toprule
\textbf{Setting} & \textbf{Content} \\
\midrule
\endfirsthead

\multicolumn{2}{c}{{\bfseries \tablename\ \thetable{} -- continued from previous page}} \\
\toprule
\textbf{Setting} & \textbf{Content} \\
\midrule
\endhead

\bottomrule
\multicolumn{2}{r}{\textit{Continued on next page...}} \\
\endfoot

\bottomrule
\endlastfoot

\multicolumn{2}{c}{\colorbox{gray!20}{\textbf{Case 1: LLaDA-8B-Instruct}}} \\
\cmidrule(lr){1-2}
\textbf{Prompt} & Ellie has found an old bicycle in a field and thinks it just needs some oil to work well again. She needs 10ml of oil to fix each wheel and will need another 5ml of oil to fix the rest of the bike. How much oil does she need in total to fix the bike? \\ \cmidrule(lr){1-2}
\textbf{LLaDA-8B-Instruct} \newline  & 
\textbf{Latency: 9.84s} \newline
\textit{Output:} To determine the total amount of oil Ellie needs to fix the bike, we can break down the problem into two parts: the oil needed for the wheels and the oil needed for the rest of the bike.

Step 1: Calculate the oil needed for the wheels. \newline
Ellie needs 10ml of oil to fix each wheel. Since a bicycle has two wheels, the total oil needed for the wheels is:
\[ 10 \text{ ml/wheel} \times 2 \text{ wheels} = 20 \text{ ml} \]

Step 2: Add the oil needed for the rest of the bike. \newline
Ellie needs an additional 5ml of oil to fix the rest of the bike. Therefore, the total amount of oil needed is:
\[ 20 \text{ ml} + 5 \text{ ml} = 25 \text{ ml} \]

So, the total amount of oil Ellie needs to fix the bike is \boxed{25} ml. \\ 
\cmidrule(lr){1-2}
\textbf{+DC-Leap} \newline  & 
\textbf{Latency: 1.95s} \newline
\textit{Output:} To determine the total amount of oil Ellie needs to fix the bike, we can break down the problem into smaller steps:

Step 1: Calculate the amount of oil needed for the wheels. \newline
Ellie needs 10ml of oil to fix each wheel. Since a bicycle has two wheels, we multiply the amount of oil needed for one wheel by 2:
\[ 10 \text{ ml/wheel} \times 2 \text{ wheels} = 20 \text{ ml} \]

Step 2: Add the amount of oil needed for the rest of the bike. \newline
Ellie needs an additional 5ml of oil to fix the rest of the bike. We add this to the amount of oil needed for the wheels:
\[ 20 \text{ ml} + 5 \text{ ml} = 25 \text{ ml} \]

Therefore, the total amount of oil Ellie needs to fix the bike is \boxed{25} ml. \\ 
\midrule[\heavyrulewidth]

\multicolumn{2}{c}{\colorbox{gray!20}{\textbf{Case 2: LLaDA-1.5}}}\\
\cmidrule(lr){1-2}
\textbf{Prompt} & Ellie has found an old bicycle in a field and thinks it just needs some oil to work well again. She needs 10ml of oil to fix each wheel and will need another 5ml of oil to fix the rest of the bike. How much oil does she need in total to fix the bike? \\ \cmidrule(lr){1-2}
\textbf{LLaDA-1.5} \newline  & 
\textbf{Latency: 9.60s} \newline
To determine the total amount of oil Ellie needs to fix the bike, we can break down the problem into two parts: the oil needed for the wheels and the oil needed for the rest of the bike.\newline\newline

Step 1: Calculate the oil needed for the wheels \newline
Ellie needs 10ml of oil for each wheel. Since a bicycle has two wheels, the total oil needed for the wheels is:
\[ 10 \text{ ml/wheel} \times 2 \text{ wheels} = 20 \text{ ml} \]

Step 2: Calculate the oil needed for the rest of the bike. \newline
Ellie needs 5ml of oil for the rest of the bike. Therefore, the total oil needed for the rest of the bike is:
\[ 5 \text{ ml} \]

Step 3: Add the oil needed for the wheels and the rest of the bike to find the total amount of oil needed.
\[ 20 \text{ ml} + 5 \text{ ml} = 25 \text{ ml} \]

So, the total amount of oil Ellie needs in total is \boxed{25} ml. \\ \cmidrule(lr){1-2}
\textbf{+DC-Leap} \newline  & 
\textbf{Latency: 1.71s} \newline
To determine the total amount of oil Ellie needs to fix the bike, we can break down the problem into smaller steps:

Step 1: Calculate the amount of oil needed for the wheels. \newline
Ellie needs 10ml of oil to fix each wheel. Assuming the bicycle has two wheels, the total amount of oil needed for the wheels is:
\[ 10 \text{ ml/wheel} \times 2 \text{ wheels} = 20 \text{ ml} \]

Step 2: Add the amount of oil needed for the rest of the bike. \newline
Ellie needs an additional 5ml of oil to fix the rest of the bike. Therefore, the total amount of oil needed is:
\[ 20 \text{ ml} + 5 \text{ ml} = 25 \text{ ml} \]

So, the total amount of oil Ellie needs to fix the bike is \boxed{25} ml. \\ 
\midrule[\heavyrulewidth]

\multicolumn{2}{c}{\colorbox{gray!20}{\textbf{Case 3: Dream-v0-7B-Instruct}}}\\
\cmidrule(lr){1-2}
\textbf{Prompt} & Ellie has found an old bicycle in a field and thinks it just needs some oil to work well again. She needs 10ml of oil to fix each wheel and will need another 5ml of oil to fix the rest of the bike. How much oil does she need in total to fix the bike? \\ \cmidrule(lr){1-2}
\textbf{Dream-v0-7B-Instruct} \newline  & 
\textbf{Latency: 8.30s} \newline
Ellie needs 10ml of oil to fix each wheel, and there are 2 wheels on a bicycle. So, she needs 10ml x 2 = 20ml of oil for the wheels.\newline\newline
Additionally, she needs 5ml of oil to fix the rest of the bike.\newline\newline
Therefore, the total amount of oil Ellie needs is 20ml + 5ml = 25ml.\newline\newline
Ellie needs \boxed{25}\$  ml of oil in total to fix the bike. \\ \cmidrule(lr){1-2}
\textbf{+DC-Leap} \newline  & 
\textbf{Latency: 1.32s} \newline
Ellie needs 10ml of oil to fix each wheel, so she needs 10ml x 2 = 20ml of oil for the wheels. \newline
She also needs 5ml of oil to fix the rest of the bike. \newline
Therefore, Ellie needs a total of 20ml + 5ml = 25ml of oil to fix the bike. \newline
\#\#\#\# 25 \newline
The answer is: 25 \\

\end{longtable}
\section{Analysis of Decoding Dynamics}
\label{sec:app_d}
\begin{figure}[h]
    \centering
    \includegraphics[width=0.45\textwidth]{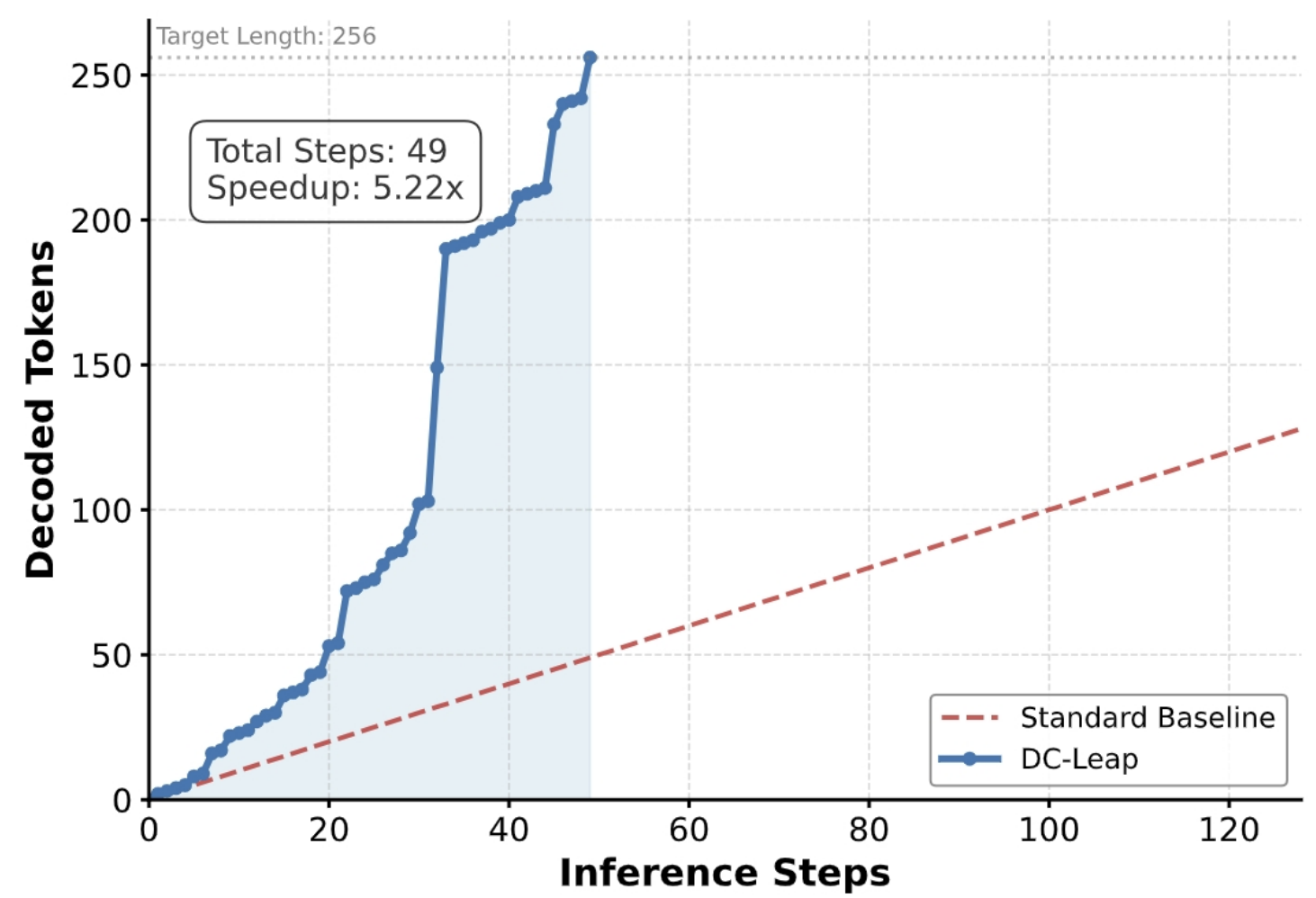} 
    \caption{\textbf{Cumulative decoded tokens with inference steps advancing.} DC-Leap (blue) shows a bursty decoding pattern, enabling massive jumps in generation progress (e.g., around step 35) compared to the linear standard baseline (red). This sample achieves a total speedup of 5.22$\times$ on a 256-token task.}
    \label{fig:decoding_efficiency}
\end{figure}
To further understand the acceleration mechanism of DC-Leap, we analyze the cumulative number of decoded tokens across inference steps. We perform this analysis on a random GSM8K sample with a generation length of 256 tokens.

Figure~\ref{fig:decoding_efficiency} visualizes the decoding trajectory of DC-Leap compared to the standard diffusion baseline. Compared to baseline, DC-Leap exhibits a non-linear, bursty generation pattern. We observe three distinct phases in the DC-Leap trajectory:
\paragraph{Exploration Phase (Steps 0-20).} In the initial steps, the slope is relatively shallow as DCV strictly improves the acceptance quality.
\paragraph{Acceleration Phase (Steps 20-35).} A nearly vertical surge in decoded tokens occurs around step 22, showcasing the synergy between drafts and the Dynamic Contiguous Verification (DCV) mechanism. In this phase, drafts serve as future anchors, progressively stabilizing the context in the decoding window. DCV capitalizes on this draft-guided coherence by applying a relaxed confidence threshold over continuous token spans. This integration allows the model to prioritize structural continuity, enabling effectively leaping through the generation process.
\paragraph{Convergence Phase (Steps 35-49).} The generation completes rapidly, achieving the target length of 256 tokens in just 49 steps.

This analysis reveals DC-Leap as a dynamic, intelligent acceleration strategy centered on Dynamic
Contiguous Verification and Draft mechanism. By enforcing a structural continuity constraint, this mechanism ensures generation quality even at lower confidence thresholds, enabling the simultaneous commitment of extended token spans. During the Acceleration Phase, this strategy effectively capitalizes on the rising confidence across the decoding window to leap through the sequence with quality drafts, facilitating convergence,  significantly accelerating through massive, verified parallel updates.

\section{Analysis of Continuous Drafts}
\label{sec:app_e}

\subsection{Motivation and Hypothesis}
\label{sec:app_e1}
In the main methodology of DC-Leap, we demonstrate that enforcing a contiguous verification constraint on the prefix allows us to safely lower the commitment threshold without compromising generation quality. Building on this, a natural hypothesis arises: \textit{Can we extend this contiguous constraint to the Draft mechanism to enable lower draft thresholds?}

Intuitively, if we enforce that drafts must also form a continuous chain starting from the verification boundary, we might be able to safely relax the draft threshold $\tau_{\text{draft}}$ significantly. Therefore, this could allow the model to leap further into the future with a denser draft sequence, potentially yielding higher speedups without compromising performance.

\subsection{Experimental Setup}
\label{sec:app_e2}
To test this hypothesis, we implement a variant of DC-Leap with Contiguous Drafts. Unlike the standard Draft mechanism, where any future token exceeding $\tau_{\text{draft}}$ is cached regardless of intermediate gaps, the Contiguous Draft strategy updates the bank if and only if:
\begin{equation}
    d_j \leftarrow \hat{x}_j \quad \text{iff} \quad \prod_{k=L_m+L_\text{prefix}}^{j} \mathbb{I}(c_{t+k} > \tau_{\text{draft}}) = 1 ,
\end{equation}
where $L_m$ and $L_\text{prefix}$ denote the maximum window size and the prefix length in DCV, respectively. And $\hat{x}_j$ represents the candidate token generated at position $j$, while $d_j$ denotes the finalized draft token accepted for the current decoding iteration. We evaluate this variant on GSM8K using LLaDA-1.5 with a fixed $\tau_{\text{commit}}=0.7$ and varying $\tau_{\text{draft}}$ from $1.00$ down to $0.00$.

\subsection{Results}
\label{sec:app_e3}
The experimental results are presented in Table~\ref{tab:continuous_draft}.

\begin{table}[h]
\centering
\caption{\textbf{Performance of Contiguous Draft Strategy on GSM8K.} We fix the commitment threshold $\tau_{\text{commit}}=0.7$ and vary the draft threshold $\tau_{\text{draft}}$.}
\label{tab:continuous_draft}
\small
\renewcommand{\arraystretch}{1.2}
\begin{tabular}{ccccc}
\toprule
\textbf{$\tau_{\text{draft}}$} & \textbf{Throughput (TPS)} & \textbf{Speedup} & \textbf{Accuracy (\%)} \\
\midrule
1.00 & 51.73 & 4.89$\times$ & 79.53 \\
0.95 & 52.10 & 4.92$\times$ & 79.53 \\
0.80 & 51.72 & 4.89$\times$ & 79.53 \\
0.70 & 51.89 & 4.90$\times$ & 79.53 \\
0.65 & 50.96 & 4.82$\times$ & 79.53 \\
0.60 & 51.97 & 4.91$\times$ & 79.53 \\
0.50 & 51.92 & 4.91$\times$ & 79.61 \\
0.40 & 51.17 & 4.84$\times$ & 78.85 \\
0.30 & 52.43 & 4.96$\times$ & 78.39 \\
0.20 & 51.71 & 4.89$\times$ & 76.19 \\
0.10 & 54.31 & 4.95$\times$ & 63.15 \\
0.00 & 62.49 & 5.91$\times$ & 59.82 \\
\bottomrule
\end{tabular}
\end{table}

\subsection{Observations and Discussions}
\label{sec:app_e4}
\paragraph{Observations.}
The results reveal three distinct behavioral phases, offering experimental evidence against the contiguous drafting mechanism:
\begin{itemize}
    \item \textbf{Stagnant Phase ($\tau_{\text{draft}} \in [0.60,1.00]$):} Accuracy remains strictly invariant at $79.53\%$, and throughput stagnates around $51 \sim 52$ TPS. This plateau suggests that the contiguous constraint severely restricts the length of generated drafts, making it virtually impossible to form a sequence that extends beyond the dynamic decoding window. Consequently, no effective drafts are cached, rendering the mechanism functionally equivalent to the baseline.
    \item \textbf{Degradation Phase ($\tau_{\text{draft}} \in [0.20, 0.60)$):} As the threshold is lowered further, accuracy begins to decline without yielding throughput gains. This suggests that while longer draft chains are finally being formed, they inevitably contain low-quality tokens due to the lower confidence threshold. These poor-quality drafts act as noise rather than informative anchors, confusing the model without facilitating convergence.
    \item \textbf{Collapse Phase ($\tau_{\text{draft}} \in [0, 0.20)$):} Only at extremely low thresholds does throughput perceptibly increase, but this comes at the cost of a catastrophic accuracy collapse. In this range, the model blindly accepts incorrect drafts, leading to rapid but nonsensical generation.
\end{itemize}
\paragraph{Discussions.}
These findings expose a fundamental conflict between the draft reliability of dLLMs and the contiguous constraint:
\begin{itemize}
    \item \textbf{The Domino Effect:} The cumulative product operation enforces a domino effect: a single low-confidence token truncates the entire future draft chain. To overcome this truncation and generate a draft long enough to extend beyond the dynamic decoding window, it becomes necessary to drastically reduce $\tau_{\text{draft}}$. However, doing so inevitably introduces erroneous tokens into the future context. Thus, contiguous drafting faces an unsolvable dilemma: the high thresholds required for guaranteeing quality and the low thresholds needed for ensuring drafts' length are mutually exclusive.
    \item \textbf{Failure to Provide Informative References:} The fundamental purpose of the draft-guided decoding strategy is to provide reliable semantic anchors that the bidirectional attention mechanism can attend to, thereby resolving uncertainty in the current masked region. However, high-confidence tokens often appear scattered in the future context. By enforcing continuity, we discard this valuable information. This confirms that the draft strategy we proposed in Section~\ref{sec:Methodology} is superior: it prioritizes the quality and informativeness of the future context over its continuity, ensuring that the drafts serve as valid references rather than noise.
\end{itemize}
\section{More Results and Analysis}
\label{sec:app_f}
\begin{table}[t]
\centering
\caption{\textbf{Ablations of commitment threshold $\tau_c$ across three models.}}
\label{tab:ablation_commit}
\begin{tabular}{llcccccc}
\toprule
\textbf{Metric} & \textbf{Model} & \textbf{0.65} & \textbf{0.68} & \textbf{0.70} & \textbf{0.72} & \textbf{0.75} & \textbf{0.80} \\ 
\midrule
Accuracy & LLaDA-8B-Instruct & 76.19 & 76.95 & 76.65 & 77.03 & 77.10 & 77.48 \\
         & LLaDA-1.5          & 79.08 & 79.98 & 80.21 & 80.59 & 81.20 & 81.20 \\
         & Dream-v0-7B-Instruct & 73.09 & 73.77 & 73.92 & 74.15 & 74.37 & 74.45 \\ 
\midrule
TPS      & LLaDA-8B-Instruct & 54.64 & 52.39 & 50.22 & 48.12 & 45.51 & 41.46 \\
         & LLaDA-1.5          & 57.84 & 54.86 & 52.07 & 50.65 & 48.13 & 44.24 \\
         & Dream-v0-7B-Instruct & 80.87 & 77.21 & 75.87 & 73.60 & 69.86 & 65.54 \\ 
\bottomrule
\end{tabular}
\end{table}
\subsection{More Results of Ablations}
\label{sec:app_f1}
In this section, we extend our sensitivity analysis to all three models: LLaDA-8B-Instruct~\citep{llada}, LLaDA-1.5~\citep{llada1.5}, and Dream-v0-7B-Instruct~\citep{dream7b}. The results (detailed in Table~\ref{tab:ablation_commit}, Table~\ref{tab:ablation_draft}, and Table~\ref{tab:ablation_window}) on the GSM8K benchmark with a fixed generation length of 256 illustrate the trade-offs between generation quality and efficiency, providing a clearer understanding of our design choices. 

\paragraph{Sensitivity to commitment threshold $\tau_c$.} Firstly, we fix $\tau_d$ = 0.98 and vary $\tau_c$ from 0.65 to 0.80. As shown in Table~\ref{tab:ablation_commit}, across all models, increasing $\tau_c$ consistently improves accuracy but leads to a reduction in TPS. Overall, accuracy remains consistently high and stable across all models, demonstrating that DC-Leap is robust to the choice of $\tau_c$ within the above range.

\begin{table}[t]
\centering
\caption{\textbf{Ablations of draft threshold $\tau_d$ across three models.}}
\label{tab:ablation_draft}
\begin{tabular}{llcccccc}
\toprule
\textbf{Metric} & \textbf{Model} & \textbf{0.85} & \textbf{0.88} & \textbf{0.90} & \textbf{0.92} & \textbf{0.95} & \textbf{0.98} \\ 
\midrule
Accuracy & LLaDA-8B-Instruct & 77.48 & 77.03 & 77.48 & 77.86 & 77.26 & 76.65 \\
         & LLaDA-1.5          & 78.77 & 79.76 & 79.68 & 80.36 & 79.98 & 80.21 \\
         & Dream-v0-7B-Instruct & 71.87 & 72.02 & 72.40 & 73.09 & 73.39 & 73.92 \\ 
\midrule
TPS      & LLaDA-8B-Instruct & 48.66 & 49.15 & 49.90 & 49.77 & 50.00 & 52.39 \\
         & LLaDA-1.5          & 57.45 & 55.96 & 54.96 & 54.89 & 54.27 & 52.07 \\
         & Dream-v0-7B-Instruct & 74.90 & 75.18 & 75.08 & 75.48 & 74.55 & 75.87 \\ 
\bottomrule
\end{tabular}
\end{table}
\paragraph{Sensitivity to draft threshold $\tau_d$.}
For draft threshold $\tau_d$, we fix the commitment threshold $\tau_c$ at 0.70 and vary $\tau_d$ from 0.65 to 0.8 $\tau_d$ across three models. From Table~\ref{tab:ablation_draft}, we can observe that LLaDA-8B-Instruct exhibits an inverse trend in both accuracy and TPS compared to LLaDA-1.5 and Dream-v0-7B-Instruct, which can be analyzed through their distinct training paradigms. 

LLaDA-8B-Instruct is trained from scratch with a mask-diffusion objective; it is inherently robust to non-contiguous and fragmented contexts. For this model, a lower $\tau_d$ aligns better with its pre-training distribution, which can facilitate faster and more reliable generation, even if individual draft tokens are less precise.

While sharing the same backbone as LLaDA-8B-Instruct, LLaDA-1.5 is refined via VRPO to pursue high-reward and deterministic outputs, making it more sensitive to the quality of right-side contexts. So a higher $\tau_d$ guarantees high-quality drafts, leading to an improvement in both accuracy and TPS.

Dream-v0-7B-Instruct inherits its weights from an autoregressive model with a strong left-to-right causal prior. This architectural heritage makes it more sensitive to low-quality non-causal drafts. 
A higher $\tau_d$ ensures a more precise right-side context, resulting in better performance.

To sum up, for LLaDA-8B-Instruct, the accuracy remains relatively stable across different $\tau_d$ with acceptable losses, whereas LLaDA-1.5 and Dream-v0-7B-Instruct are more sensitive to this parameter. Therefore, $\tau_d$ = 0.98 serves as an ideal balance that works well for all these models. In addition, all three models exhibit relatively stable performance in both accuracy and TPS.

\begin{table}[t]
\centering
\caption{\textbf{Ablations of maximum window size $L$ across three models.}}
\label{tab:ablation_window}
\begin{tabular}{llcccc}
\toprule
\textbf{Metric} & \textbf{Model} & \textbf{32} & \textbf{64} & \textbf{96} & \textbf{128} \\ 
\midrule
Accuracy & LLaDA-8B-Instruct    & 76.12 & 76.57 & 76.65 & 76.65 \\
         & LLaDA-1.5            & 80.59 & 80.74 & 80.29 & 80.21 \\
         & Dream-v0-7B-Instruct & 74.15 & 74.07 & 74.15 & 73.92 \\ 
\midrule
TPS      & LLaDA-8B-Instruct    & 49.26 & 49.91 & 49.70 & 52.39 \\
         & LLaDA-1.5            & 29.60 & 51.92 & 51.08 & 52.07 \\
         & Dream-v0-7B-Instruct & 70.34 & 74.22 & 75.16 & 75.87 \\ 
\bottomrule
\end{tabular}
\end{table}

\paragraph{Sensitivity to maximum window length $L$:}
For the maximum window size $L$, we fix $\tau_c$ at 0.70 and $\tau_d$ at 0.98, and Table~\ref{tab:ablation_window} records the accuracy and TPS in different $\tau_c$ in three models. In general, all of three models show high stability regarding the window size $L$. As L increases from 32 to 128, the accuracy remains steady, while the TPS shows a relatively clear gain. This confirms that our method's efficiency scales well with the window size across different architectures.  

Actually, the observed stability in accuracy and the increase in TPS as $L$ expands are consistent with the design intuition of our Dynamic Contiguous Verification (DCV) mechanism. By enforcing a strictly sequential and contiguous verification, DCV ensures that only parallel tokens satisfying the sequential chain are accepted. This decouples generation quality from the maximum window size $L$ to some extent, since accuracy is the same once $L$ is large enough to cover all tokens that pass the DCV. This also explains why the actual speedup exhibits diminishing marginal improvements as $L$ increases.

\subsection{More Results of Step Distilled dLLMs}
\label{sec:app_f2}
\begin{table}[t]
\centering
\caption{\textbf{Performance on dParallel-LLaDA-8B-instruct~\citep{dparallel}.}}
\label{tab:results_dparallel1}
\begin{tabular}{llcccc}
\toprule
\textbf{Model} & \textbf{Metric} & \textbf{\shortstack{GSM8K-CoT\\(0 shot)}} & \textbf{\shortstack{HumanEval\\(0 shot)}} & \textbf{\shortstack{MATH\\(4 shot)}} & \textbf{\shortstack{MBPP\\(3 shot)}} \\ 
\midrule
dParallel-LLaDA-8B-instruct & Accuracy & 75.06 & 39.00 & 35.84 & 39.40 \\
                            & TPS      & 186.11 (1.0$\times$) & 76.38 (1.0$\times$) & 65.23 (1.0$\times$) & 120.53 (1.0$\times$) \\
\midrule
+DC-Leap                    & Accuracy & 74.75 & 37.42 & 34.48 & 41.20 \\
                            & TPS      & 256.54 (1.38$\times$) & 141.60 (1.85$\times$) & 95.70 (1.47$\times$) & 290.10 (2.40$\times$) \\
\bottomrule
\end{tabular}
\end{table}

\begin{table}[t]
\centering
\caption{\textbf{Performance on dParallel-Dream-7B-Instruct~\citep{dparallel}.}}
\label{tab:results_dparallel2}
\begin{tabular}{llcccc}
\toprule
\textbf{Model} & \textbf{Metric} & \textbf{\shortstack{GSM8K-CoT\\(0 shot)}} & \textbf{\shortstack{HumanEval-Instruct\\(0 shot)}} & \textbf{\shortstack{MATH\\(0 shot)}} & \textbf{\shortstack{MBPP-Instruct\\(0 shot)}} \\ 
\midrule
dParallel-Dream-7B-Instruct & Accuracy & 82.94 & 54.88 & 39.88 & 55.20 \\
                            & TPS      & 190.16 (1.0$\times$) & 127.03 (1.0$\times$) & 70.61 (1.0$\times$) & 241.51 (1.0$\times$) \\
\midrule
+DC-Leap                    & Accuracy & 80.52 & 53.05 & 37.48 & 54.00 \\
                            & TPS      & 257.28 (1.35$\times$) & 206.27 (1.62$\times$) & 90.32 (1.28$\times$) & 256.54 (1.06$\times$) \\
\bottomrule
\end{tabular}
\end{table}
\begin{table}[h]
  \centering
  \caption{\textbf{Performance comparison on LLaDA-8B-Instruct using three confidence definitions.} We compare the Top-1 Confidence baseline with Calibrated Confidence and Semantic Entropy. Accuracy (\%) and TPS are reported.}
  \label{tab:conf_llada}
  \small
  \setlength{\tabcolsep}{10pt}
  \begin{tabular}{llcccc}
    \toprule
    \textbf{Method} & \textbf{Metric} & \textbf{GSM8K} & \textbf{MBPP} & \textbf{HumanEval} & \textbf{IFEval} \\
    \midrule
    
    \multirow{2}{*}{Top-1} & Accuracy & 76.65 & 29.60 & 35.37 & 71.34 \\
     & TPS & 50.22 (4.80$\times$) & 59.30 (4.58$\times$) & 90.72 (4.73$\times$) & 72.48 (3.44$\times$) \\
    
    \midrule
    
    \multirow{2}{*}{Calibrated Confidence} & Accuracy & 77.48 & 29.80 & 35.37 & 70.14 \\
     & TPS & 25.04 (2.39$\times$) & 30.42 (2.35$\times$) & 40.19 (2.10$\times$) & 45.61 (2.16$\times$) \\
    
    \midrule
    
    \multirow{2}{*}{Semantic Entropy} & Accuracy & 77.41 & 30.0 & 36.59 & 71.34 \\
     & TPS & 28.97 (2.77$\times$) & 32.28 (2.49$\times$) & 32.84 (1.71$\times$) & 24.76 (1.18$\times$) \\
     
    \bottomrule
  \end{tabular}
\end{table}
\begin{table}[h]
  \centering
  \caption{\textbf{Performance comparison on LLaDA-1.5 using three confidence definitions.} We compare the Top-1 Confidence baseline with Calibrated Confidence and Semantic Entropy. Accuracy (\%), TPS are reported.}
  \label{tab:conf_llada1.5}
  \small
  \setlength{\tabcolsep}{10pt}
  \begin{tabular}{llcccc}
    \toprule
    \textbf{Method} & \textbf{Metric} & \textbf{GSM8K} & \textbf{MBPP} & \textbf{HumanEval} & \textbf{IFEval} \\
    \midrule
    
    \multirow{2}{*}{Top-1} & Accuracy & 80.21 & 42.60 & 38.41 & 71.82 \\
     & TPS & 52.07 (4.92$\times$) & 171.35 (13.04$\times$) & 102.74 (5.30$\times$) & 69.67 (3.22$\times$) \\
    
    \midrule
    
    \multirow{2}{*}{Calibrated Confidence} & Accuracy & 81.58 & 40.40 & 40.24 & 70.62 \\
     & TPS & 26.48 (2.50$\times$) & 49.22 (3.76$\times$) & 40.63 (2.10$\times$) & 40.84 (1.89$\times$) \\
    
    \midrule
    
    \multirow{2}{*}{Semantic Entropy} & Accuracy & 80.06 & 43.00 & 39.02 & 72.30 \\
     & TPS & 26.65 (2.52$\times$) & 25.99 (1.98$\times$) & 34.31 (1.77$\times$) & 24.80 (1.15$\times$) \\
     
    \bottomrule
  \end{tabular}
\end{table}
dParallel~\citep{dparallel} is a distilled dLLM specifically optimized to enhance the certainty of parallel token prediction, thereby unlocking the acceleration potential of generation. By leveraging a significantly lower confidence threshold than conventional parallel decoding, typically 0.5 in most benchmarks and 0.45 in others, dParallel achieves substantial inference speedups without compromising generative quality.

All experiments are conducted on NVIDIA RTX 6000 Blackwell GPUs. We strictly adhere to the original configurations of dParallel for reproduction. For the integration of DC-Leap, we utilize a consistent hyperparameter set across both dParallel-LLaDA-8B-Instruct and dParallel-Dream-7B-Instruct: $\tau_c$ = 0.75, $\tau_d$ = 0.98, and maximum window size $L$ = 128. The performance, throughput (TPS), and relative speedups are summarized in the Table~\ref{tab:results_dparallel1} and Table~\ref{tab:results_dparallel2}.

It is worth noting that since dParallel already minimizes sampling steps to a highly compressed range for most inputs, further step reduction typically leads to performance degradation. However, DC-Leap demonstrates its robustness in this constrained scenario, successfully delivering additional throughput gains with comparable accuracy.

\subsection{Analysis of Alternative Confidence Definition}
\label{sec:app_f3}

In this section, we consider alternative confidence definitions beyond Top-1 probability, including calibrated confidence~\citep{calibrated}, entropy-based uncertainty~\citep{semantic} and self-assessed confidence~\citep{ioe}. Specifically, we conduct an analysis integrating the three confidence definitions mentioned above into DC-Leap across all three models. The results are summarized in Table~\ref{tab:conf_llada}, Table~\ref{tab:conf_llada1.5}, Table~\ref{tab:conf_dream} and Table~\ref{tab:ioe_conf}. Our results indicate that DC-Leap benefits from the Top-1 confidence formulation, as it demonstrates the strongest synergy with our DCV mechanism.

\paragraph{Analysis of Calibrated Confidence and Semantic Entropy.} We evaluate these as alternative confidence estimators by substituting our Top-1 strategy across three models and four benchmarks. As shown in Table~\ref{tab:conf_llada}, Table~\ref{tab:conf_llada1.5} and Table~\ref{tab:conf_dream}, although both alternatives yield marginal accuracy improvements over Top-1, they significantly reduce the achieved speedup.

We attribute this to the strictness of our DCV mechanism: these more sophisticated estimators impose tighter threshold criteria, reducing the number of tokens committed per forward pass. This suggests that an ideal confidence definition for DC-Leap should not only enhance individual token acceptance accuracy but also remain flexible enough to permit acceptance of more tokens in sequence. Currently, the Top-1 confidence formulation balances these two objectives.

\paragraph{Self-Assessed Confidence.} Regarding self-assessed confidence, since the IoE-prompt design is specifically tailored for tasks with deterministic answers rather than open-ended generation, we conduct evaluations on the GSM8K benchmark using LLaDA-8B-Instruct and LLaDA-1.5. The results are detailed in Table~\ref{tab:ioe_conf}.

Initially, we maintain the original hyperparameters of  $\tau_c$ = 0.7 and $\tau_d$ = 0.98 to ensure a fair comparison with the DC-Leap baseline. We then integrate the IoE-prompt to enable model self-assessment. The results include accuracy, TPS (with speedup ratio), and total runtime seconds, which are summarized in the table below.

As observed, self-assessed confidence is effective, leading to an improvement in both accuracy and TPS. This gain is likely because the self-assessment process enhances the determinism of the preceding context. Consequently, during the subsequent self-evaluation and answer verification phases, a single forward pass can process and verify a larger number of tokens. However, while the overall TPS improves a lot, the total inference time also increases due to the additional reflection process.

Furthermore, since the IoE-prompt inherently provides higher results accuracy, we explore the feasibility of relaxing the confidence threshold. Experimental evidence suggests that $\tau_c$ can be safely reduced to 0.6 without compromising accuracy. Under this configuration, the models achieve even higher TPS while remaining competitive in accuracy compared to the Top-1 baseline.

\begin{table}[h]
  \centering
  \caption{\textbf{Performance comparison on Dream-v0-7B-Instruct using three confidence definitions.} We compare the Top-1 Confidence baseline with Calibrated Confidence and Semantic Entropy. Accuracy (\%), TPS are reported.}
  \label{tab:conf_dream}
  \small
  \setlength{\tabcolsep}{10pt}
  \begin{tabular}{llcccc}
    \toprule
    \textbf{Method} & \textbf{Metric} & \textbf{GSM8K} & \textbf{MBPP} & \textbf{HumanEval} & \textbf{IFEval} \\
    \midrule
    
    \multirow{2}{*}{Top-1} & Accuracy & 73.92 & 57.60 & 59.76 & 51.92 \\
     & TPS & 75.87 (5.45$\times$) & 88.59 (5.10$\times$) & 116.21 (5.76$\times$) & 110.64 (4.06$\times$) \\
    
    \midrule
    
    \multirow{2}{*}{Calibrated Confidence} & Accuracy & 74.45 & 57.80 & 59.76 & 50.00 \\
     & TPS & 50.23 (3.61$\times$) & 59.46 (3.43$\times$) & 50.12 (2.48$\times$) & 42.32 (1.55$\times$) \\
    
    \midrule
    
    \multirow{2}{*}{Semantic Entropy} & Accuracy & 74.30 & 58.40 & 60.36 & 51.92 \\
     & TPS & 46.72 (3.36$\times$) & 23.30 (1.34$\times$) & 36.08 (1.79$\times$) & 38.90 (1.43$\times$) \\
     
    \bottomrule
  \end{tabular}
\end{table}
\begin{table}[!htbp]
  \centering
  \caption{\textbf{Evaluation of Self-Assessed Confidence on GSM8K.} We compare the Top-1 Confidence baseline with Self-assessment confidence at different confidence thresholds ($\tau_c$). Accuracy (\%), TPS, and total Runtime (s) are reported.}
  \label{tab:ioe_conf}
  \small
  \setlength{\tabcolsep}{12pt}
  \begin{tabular}{llcc}
    \toprule
    \textbf{Method} & \textbf{Metric} & \textbf{LLaDA-8B-Instruct} & \textbf{LLaDA-1.5} \\
    \midrule
    
    \multirow{3}{*}{Top-1} & Accuracy & 76.65 & 80.21 \\
     & TPS & 50.22 (4.80$\times$) & 52.07 (4.92$\times$) \\
     & Runtime/s & 6724.21 & 6484.92 \\
    
    \midrule
    
    \multirow{3}{*}{Self-Assessed ($\tau_c = 0.7$)} & Accuracy & 79.53 & 81.05 \\
     & TPS & 73.50 (7.02$\times$) & 74.99 (7.09$\times$) \\
     & Runtime/s & 9959.15 & 9844.73 \\
    
    \midrule
    
    \multirow{3}{*}{Self-Assessed ($\tau_c = 0.6$)} & Accuracy & 77.79 & 79.98 \\
     & TPS & 89.32 (8.53$\times$) & 90.87 (8.59$\times$) \\
     & Runtime/s & 8167.90 & 8081.51 \\
     
    \bottomrule
  \end{tabular}
\end{table}

\end{document}